\documentclass{article}

    \PassOptionsToPackage{numbers, compress}{natbib}
 \usepackage[preprint]{neurips_2026}

\usepackage[utf8]{inputenc} % allow utf-8 input
\usepackage[T1]{fontenc}    % use 8-bit T1 fonts
\usepackage{hyperref}       % hyperlinks
\usepackage{url}            % simple URL typesetting
\usepackage{booktabs}       % professional-quality tables
\usepackage{amsfonts}       % blackboard math symbols
\usepackage{nicefrac}       % compact symbols for 1/2, etc.
\usepackage{microtype}      % microtypography
\usepackage{xcolor}         % colors

\usepackage{graphicx}

\usepackage[table]{xcolor}
\usepackage{pifont}

\newcommand{\cmark}{\textcolor{green!60!black}{\ding{51}}}
\newcommand{\xmark}{\textcolor{red!70!black}{\ding{55}}}

\usepackage{algorithm}
\usepackage{algorithmic}
\usepackage{multirow}
\usepackage{makecell}

\usepackage{newtxtext,newtxmath}
\usepackage{bbm}  % For \mathbbm{1}

\title{\textit{RSRCC}: A Remote Sensing Regional Change Comprehension Benchmark Constructed via Retrieval-Augmented Best-of-$N$ Ranking}

\author{%
  Roie Kazoom, Yotam Gigi, George Leifman, \\
  \textbf{Tomer Shekel, Genady Beryozkin} \\
  Google Research \\
  \texttt{{kazoomroie,yotamg,gleifman,shekelto,genady} @google.com} \\
}

\begin{document}

\maketitle

\vspace{-23pt}

\begin{figure*}[!h]
  \centering
  \includegraphics[width=\textwidth]{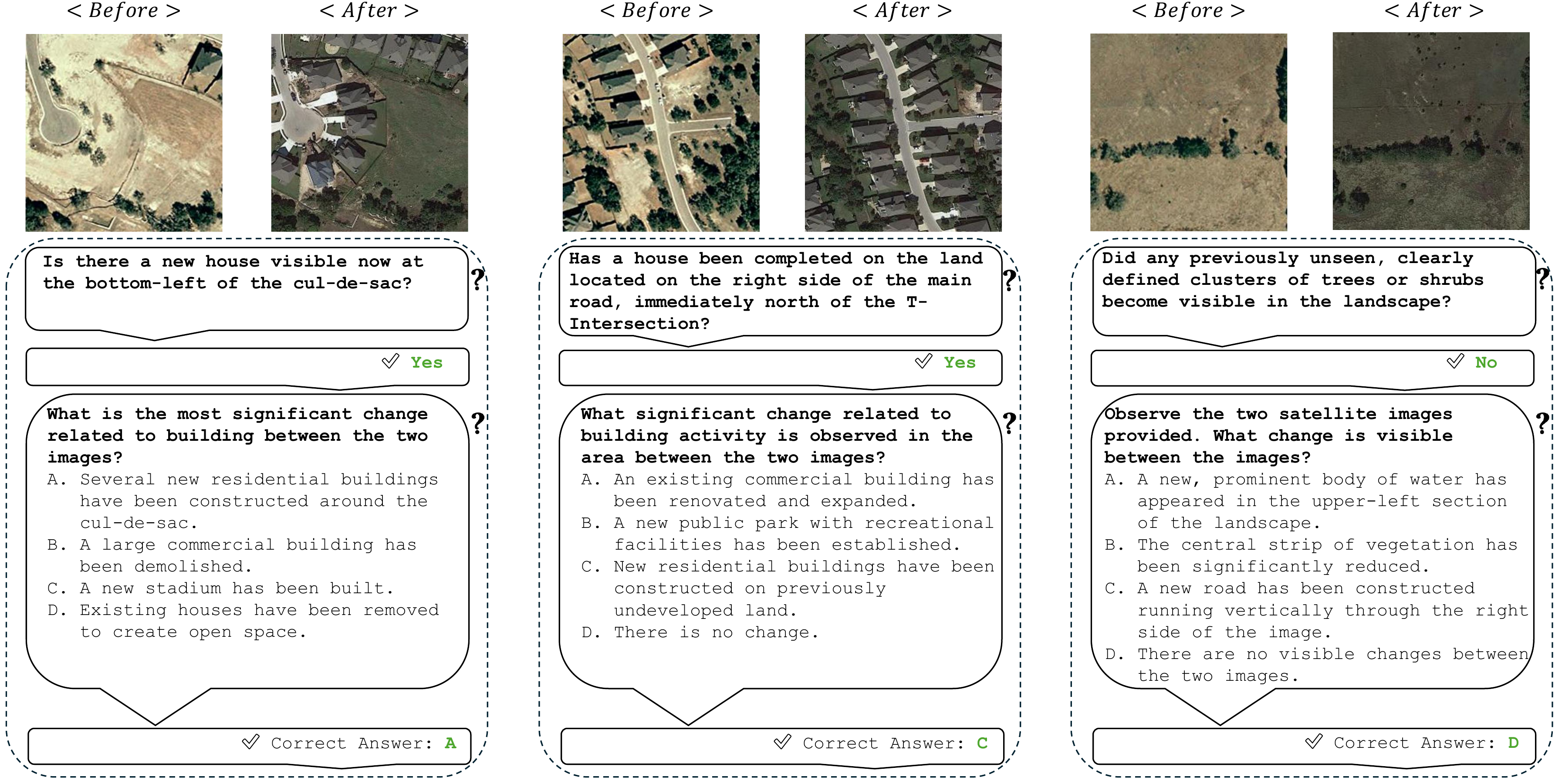}
\caption{\textbf{Example samples from our generated dataset.} Each sample contains a pair of \textit{before} and \textit{after} satellite images, followed by one or more visual localized questions. The questions are designed to capture semantic changes, such as new construction, demolition, vegetation loss, or no visible change. The correct answers are highlighted in \textcolor{green!60!black}{green} within the figure.}
    \label{fig:data_example}
\end{figure*}

\begin{abstract}
Traditional change detection identifies where changes occur, but does not explain what changed in natural language. Existing remote sensing change captioning datasets typically describe overall image-level differences, leaving fine-grained localized semantic reasoning largely unexplored. To close this gap, we present \textit{RSRCC}, a new benchmark for remote sensing change question-answering containing 126k questions, split into $87k$ training, $17.1k$ validation, and $22k$ test instances. Unlike prior datasets, \textit{RSRCC} is built around localized, change-specific questions that require reasoning about a particular semantic change. To the best of our knowledge, this is the first remote sensing change question-answering benchmark designed explicitly for such fine-grained reasoning-based supervision. To construct \textit{RSRCC}, we introduce a hierarchical semi-supervised curation pipeline that uses Best-of-$N$ ranking as a critical final ambiguity-resolution stage. First, candidate change regions are extracted from semantic segmentation masks, then initially screened using an image-text embedding model, and finally validated through retrieval-augmented vision-language curation with Best-of-$N$ ranking. This process enables scalable filtering of noisy and ambiguous candidates while preserving semantically meaningful changes. The dataset is available at \url{https://huggingface.co/datasets/google/RSRCC}.
\end{abstract}

\section{Introduction}

\begin{figure*}[t]
    \centering
    \includegraphics[width=\textwidth]{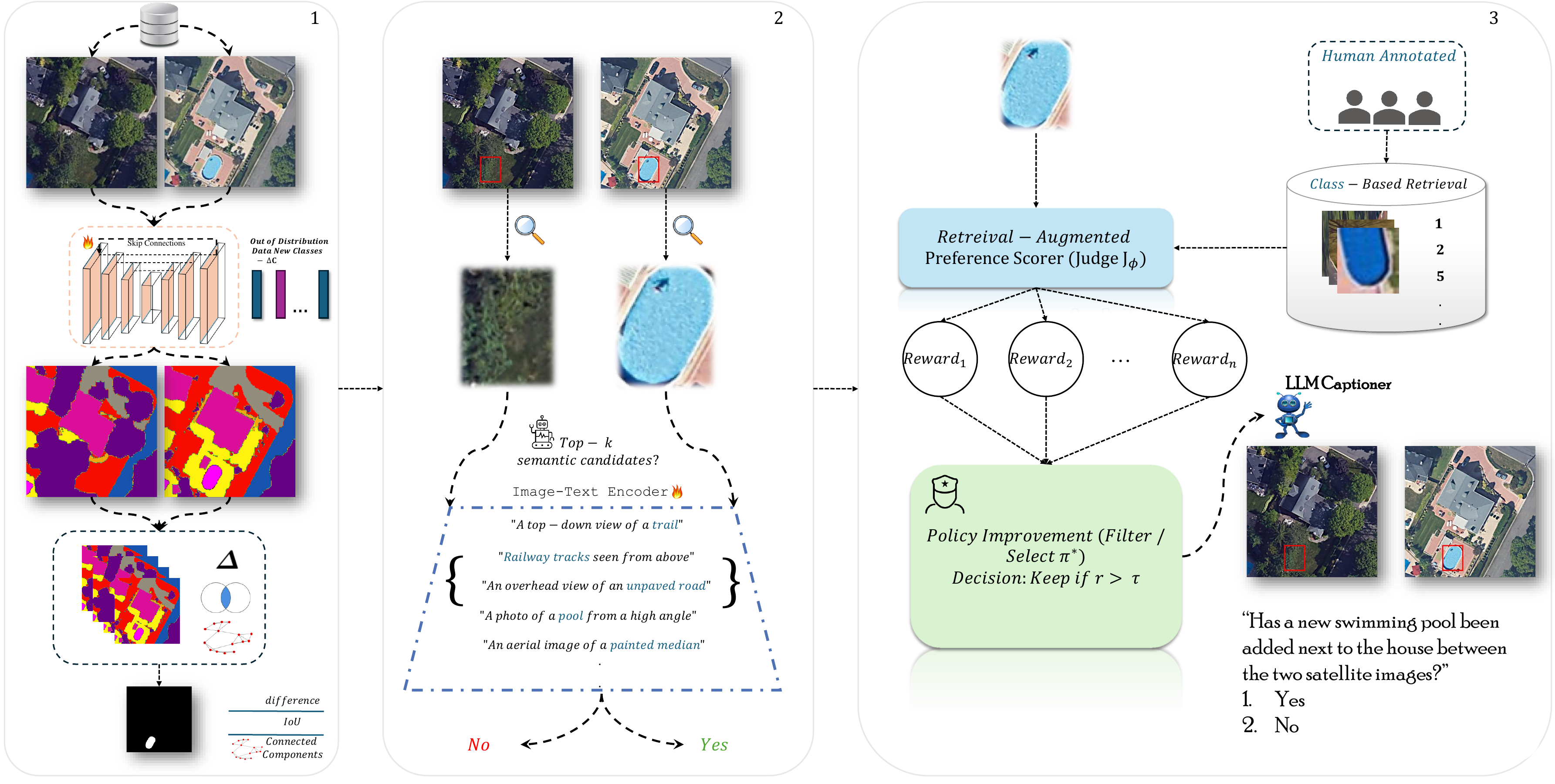}
    \caption{
        \textbf{Pipeline overview.}
        On the \textbf{left} (1), semantic mask differences are analyzed using Intersection over Union (IoU) and connected component analysis to localize candidate changes, denoted by $\Delta x$.
        In the \textbf{middle} (2), an image-text encoder processes cropped regions $x \in \mathbb{R}^{H \times W}$ and retrieves the top-$k$ most similar candidates $\{x_1, x_2, \ldots, x_k\}$ for preliminary semantic validation.
        On the \textbf{right} (3), ambiguous cases are resolved through retrieval-augmented Best-of-$N$ ranking.
        A large language model scores the query against retrieved annotated examples, assigning a relevance score $r \in \{1,..,n\}$ that promotes semantically consistent matches and filters low-confidence candidates.
    }
    \label{fig:pipeline_overview}
\end{figure*}

Change detection (CD) from satellite imagery aims to identify differences between multi-temporal scenes, supporting applications such as urban monitoring, environmental assessment, and disaster response. Traditional CD mainly focuses on \emph{where} change occurs, typically through pixel-level or object-level predictions. Over the past decade, progress has been driven by benchmark datasets such as~\citep{chen2020levircd,liu2024levirmci,liu2022levircc,shen2021s2looking,ji2023changenet,yang2021second,van2021spacenet7,zhang2020review,zhu2017deep} and architectural advances ranging from Siamese convolutional networks to transformer-based models such as \textit{ChangeFormer}~\citep{bandara2022changeformer} and \textit{TSDT}~\citep{liu2022levircc}. Despite these advances, obtaining reliable semantic supervision remains a major bottleneck, as fine-grained annotation of change types and descriptions is costly, slow, and difficult to scale. This limitation has motivated the emerging task of \emph{change captioning}, which aims to explain \emph{what} changed using natural language. Existing remote sensing change captioning datasets, however, typically describe the overall differences between two images at the scene level~\citep{liu2022levircc,chen2025rscc}. Such formulations are valuable, but they do not directly require fine-grained reasoning about a specific localized change. In many realistic settings, a model must determine whether a particular region changed, identify the semantic category of that change, and distinguish meaningful change from distractors or no-change regions. This requires region-grounded reasoning rather than only global scene summarization. 

To address this gap, we introduce \textit{RSRCC}, a new benchmark for remote sensing change question answering built around localized, change-specific questions. Rather than asking for a single caption summarizing the entire image pair, \textit{RSRCC} evaluates whether a model can reason about a particular semantic change instance. To the best of our knowledge, this is the first remote sensing change question answering benchmark designed explicitly for such fine-grained reasoning-based supervision. Figure~\ref{fig:data_example} shows representative examples from the dataset. To construct \textit{RSRCC}, we propose a hierarchical semi-supervised curation pipeline centered on \textbf{Best-of-$N$ ranking}. Candidate change regions are first extracted from semantic segmentation masks using connected component analysis. They are then validated in two stages: fast semantic screening with a remote-sensing-tuned image-text encoder, followed by retrieval-augmented semantic verification with \textbf{Best-of-$N$ ranking} for ambiguous cases. This design filters false and ambiguous changes, reducing manual annotation and improving consistency. \textbf{Our main contributions are as follows:}

\noindent\textbf{A new benchmark for localized semantic change.}
We introduce \textit{RSRCC}, a high-quality remote sensing change question-answering benchmark built from LEVIR-CD~\citep{chen2020levircd}. Unlike prior datasets that emphasize global image-pair descriptions, \textit{RSRCC} is constructed around localized, change-specific questions, with diverse semantic labels and visual patterns for training and evaluation of multimodal models.

\noindent\textbf{A hierarchical semi-supervised curation pipeline with Best-of-$N$ ranking.}
We propose a scalable data construction framework that combines connected-component-based candidate extraction, image-text semantic screening, and retrieval-augmented \textbf{Best-of-$N$ ranking} for ambiguity-aware validation.

\noindent\textbf{A theoretical interpretation of retrieval-guided validation.}
We provide an analysis showing how in-distribution retrieval and group-restricted conditioning can reduce reasoning bias by preserving local decision boundaries and limiting exposure to out-of-distribution examples (see Appendix~\ref{sec:group_restricted_retrieval}).

Table~\ref{tab:dataset_comparison} compares \textit{RSRCC} with existing remote sensing text-image datasets. While several prior datasets provide natural-language descriptions and some include bi-temporal imagery, they typically describe overall scene-level content or global before-after differences. In contrast, \textit{RSRCC} is built around localized, change-specific supervision, where each example targets a particular semantic change instance. This makes \textit{RSRCC}, to the best of our knowledge, the first benchmark in this space designed explicitly for fine-grained localized change.

\begin{table*}[!ht]
\small
\centering
\caption{Comparison with existing remote sensing text-image datasets.}
\label{tab:dataset_comparison}
\resizebox{\textwidth}{!}{%
\begin{tabular}{l c c c c c}
\toprule
\textbf{Dataset} & \textbf{Year} & \textbf{\#Captions} & \textbf{Details} & \textbf{Temporal} & \textbf{Localized Change} \\
\midrule
UCM-Captions~\citep{luo2017exploring}    & 2016 & 10,500  & \xmark & \xmark & \xmark \\
RSICD~\citep{luo2017exploring}           & 2018 & 54,605  & \xmark & \xmark & \xmark \\
fMoW~\citep{christie2018functional}            & 2018 & N/A      & \xmark & \cmark & \xmark \\
SpaceNet 7~\citep{van2021spacenet}      & 2021 & N/A      & \xmark & \cmark & \xmark \\
S2Looking~\citep{shen2021s2looking}       & 2021 & N/A      & \xmark & \cmark & \xmark \\
QFabric~\citep{verma2021qfabric}         & 2021 & N/A      & \xmark & \cmark & \xmark \\
SpaceNet 8~\citep{hansch2022spacenet}      & 2022 & N/A      & \xmark & \cmark & \xmark \\
LEVIR-CC~\citep{liu2022levircc} & 2022 & 50,385 & \cmark & \cmark & \xmark \\
Dubai-CCD~\citep{hoxha2022change}       & 2022 & 2,500    & \cmark & \cmark & \xmark \\
RSICap~\citep{hu2025rsgpt}          & 2023 & 2,585    & \cmark & \xmark & \xmark \\
RS5M~\citep{rssm2025}            & 2024 & 5M       & \cmark & \xmark & \xmark \\
VRSBench~\citep{li2024vrsbench}        & 2024 & 29,614   & \cmark & \xmark & \xmark \\
WHU-CDC~\citep{whucdc2024}         & 2024 & 37,170   & \cmark & \cmark & \xmark \\
XLRS-Bench~\citep{xlrsbench2025}      & 2025 & 934      & \cmark & \xmark & \xmark \\
RSCC~\citep{chen2025rscc} & 2025 & 62,351 & \cmark & \cmark & \xmark \\
\midrule
\rowcolor{green!10}
\textbf{RSRCC (Ours)} & \textbf{2026} & \textbf{126k} & \cmark & \cmark & \cmark \\
\bottomrule
\end{tabular}%
}
\end{table*}

\section{Related Work and Notations}

\subsection{Change Detection}

CD identifies differences between multi-temporal remote sensing images. Formally, given two co-registered images 
\begin{equation}
  I_t, I_{t+\Delta t} \in \mathbb{R}^{H \times W \times C}  
\end{equation}
acquired at times $t$ and $t+\Delta t$, the goal is to estimate a change mask 
\begin{equation}
    M = \mathcal{F}(I_t, I_{t+\Delta t}),
\end{equation}
where $M \in \{0,1\}^{H \times W}$ and $M_{ij}=1$ indicates that pixel $(i,j)$ has changed. Progress in CD has been strongly shaped by benchmark datasets. \textit{LEVIR-CD}~\citep{chen2020levircd} remains one of the most widely used resources for high-resolution building change detection. \textit{LEVIR-MCI}~\citep{liu2024levirmci} extends this setting toward finer-grained semantic classes, while \textit{LEVIR-CC}~\citep{liu2022levircc} introduces natural-language descriptions of visual changes. More recently, \textit{RSCC}~\citep{chen2025rscc} expands change captioning to disaster scenarios. Other benchmarks, such as \textit{S2Looking}~\citep{shen2021s2looking}, \textit{ChangeNet}~\citep{ji2023changenet}, \textit{SECOND}~\citep{yang2021second}, and \textit{SpaceNet 7}~\citep{van2021spacenet7}, broaden the field to more challenging viewpoints, temporal asymmetry, and diverse scenes. Surveys such as~\citep{zhang2020review,zhu2017deep} summarize these developments. Early deep learning approaches formulated CD as dense pixel-wise prediction from paired images. Siamese convolutional networks compared features extracted from the two temporal inputs using $f_{\text{diff}} = \|\phi(I_t) - \phi(I_{t+\Delta t})\|_2$, where $\phi(\cdot)$ denotes a shared encoder, and decoded this representation into a binary or semantic change mask. Later work improved temporal reasoning through attention and transformer-based architectures. ChangeStar~\citep{zheng2021changestar} introduced attention mechanisms to emphasize informative spatial regions, while \textit{ChangeFormer}~\citep{bandara2022changeformer} reformulated CD with transformer-based sequence modeling and global context aggregation. Dual-stream transformers such as \textit{DST}~\citep{liu2022levircc} further improved cross-scale temporal modeling.

\subsection{Segmentation Masks for Change Detection}

Segmentation has long been the dominant paradigm for change detection, where the task is formulated as predicting a dense pixel-wise mask indicating binary or semantic change. Given two images 
\begin{equation}
  I_t, I_{t+\Delta t} \in \mathbb{R}^{H \times W \times C},
\end{equation}
the model outputs a segmentation map  $M = \mathcal{F}(I_t, I_{t+\Delta t}),$ where $M \in \{0,1,\ldots,K\}^{H \times W}$, $K=1$ corresponds to binary CD, and $K>1$ to multi-class CD. Early CD methods relied on spectral differencing, principal component analysis, and thresholding over indices such as NDVI, producing coarse binary masks. These approaches were sensitive to illumination changes, seasonality, and registration errors. Deep learning reshaped CD by learning temporal correspondences directly from paired images. Siamese convolutional networks applied a shared encoder $\phi(\cdot)$ to both temporal inputs and compared their features through $f_{\text{diff}} = \|\phi(I_t) - \phi(I_{t+\Delta t})\|_p$, with $p \in \{1,2\}$, followed by a decoder that predicts the change mask. Later work incorporated attention mechanisms~\citep{zheng2021changestar} and transformer-based architectures~\citep{bandara2022changeformer} to model long-range spatial and temporal dependencies, while multi-scale fusion improved robustness across object sizes. Despite their strong localization ability, segmentation-based approaches remain limited for semantic change understanding. First, they produce machine-readable masks, but not human-readable explanations of what changed. At the same time, free-form language alone is not sufficient for scalable evaluation unless it is grounded in structured outputs. Our goal is therefore not to replace machine-readable supervision with text, but to complement localization with region-grounded, semantically structured language. Second, dense mask annotation is costly and difficult to scale. Datasets such as \textit{LEVIR}~\citep{chen2020levircd} and \textit{S2Looking}~\citep{shen2021s2looking} illustrate both the strengths and the limits of this paradigm: they provide high-quality change masks, but offer only limited semantic supervision. Segmentation serves as the first stage of dataset construction by localizing candidate change regions, which are then further validated for semantic consistency.

\begin{figure*}[t]
    \centering
    \includegraphics[width=0.9\textwidth]{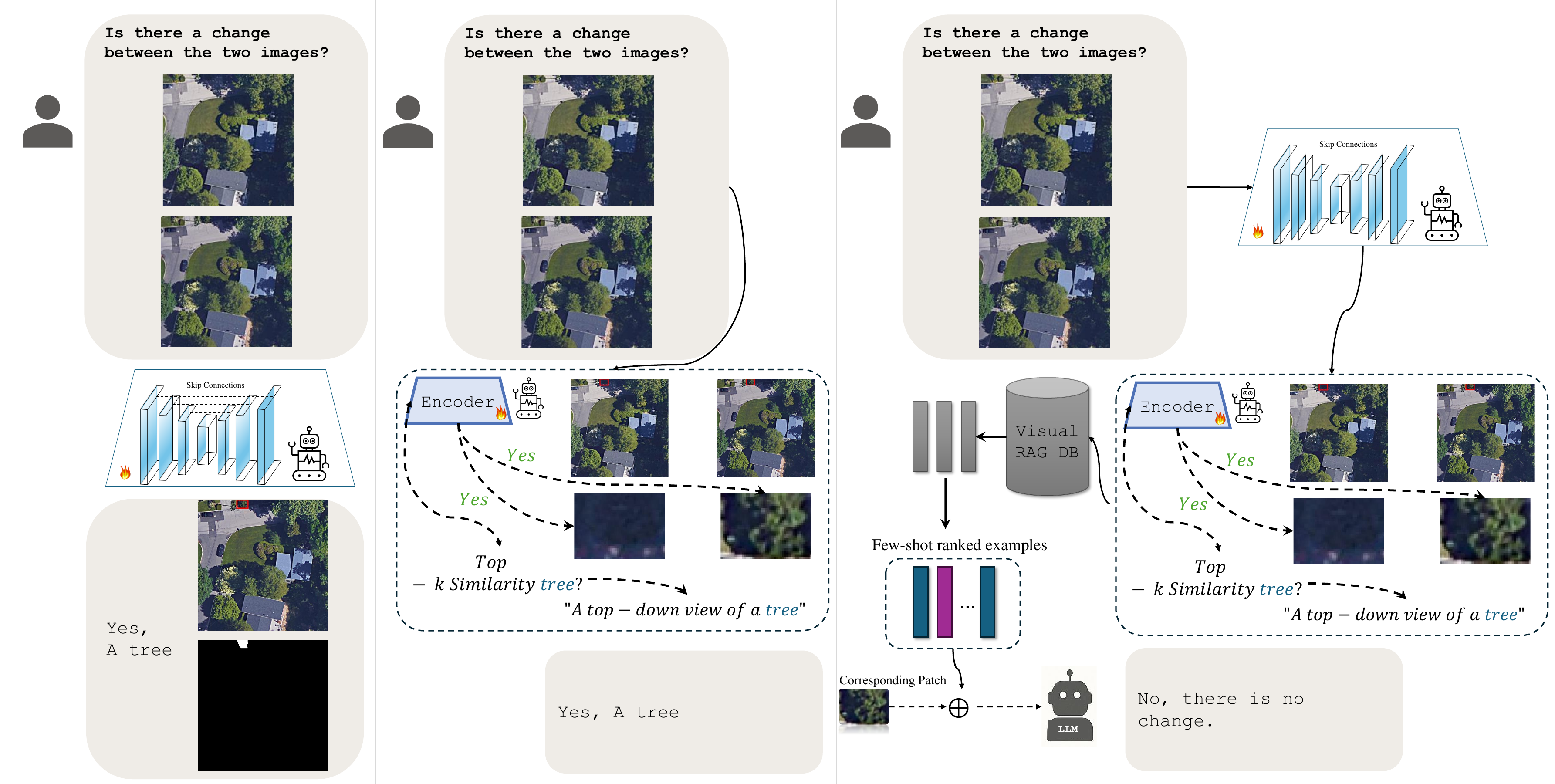}
    \caption{
    \textbf{Filtering false positives of ``no change.''}
    Segmentation models can mistakenly mark unchanged regions as changes.
    Given image patches $x_{\text{before}}$ and $x_{\text{after}}$, we use an encoder $f(\cdot)$ to compute a similarity score.
    If the similarity satisfies $s \geq \tau_{\text{sim}}$, the region is treated as a potential no-change case.
    To improve robustness, we retrieve class-based examples $\mathcal{E} = \{(x_k, y_k)\}_{k=1}^K$ and validate them with a large language model judge $J_\phi$.
    Only when $J_\phi$ confirms semantic consistency across time is the patch discarded as a false positive.
    }
    \label{fig:no_change_filter}
\end{figure*}

\subsection{Vision-Language Encoders}

Recent vision-language pretraining has produced models that align images and text in a shared embedding space, enabling semantic comparison beyond mask prediction. CLIP~\citep{radford2021clip} learns joint visual-textual representations through contrastive learning, establishing a strong foundation for multimodal alignment. SigLIP~\citep{zhai2023siglip} improves this formulation with a sigmoid-based objective, while MaMMUT~\citep{singh2022mammut} strengthens fine-grained alignment through masked image-text modeling. Several remote-sensing adaptations further demonstrate the value of vision-language pretraining for geospatial understanding, including RSCLIP~\citep{li2023rs}, RemoteCLIP~\citep{liu2023remoteclip}, and GeoCLIP~\citep{cepeda2023geoclip}. These models provide transferable semantic representations that are useful for validating whether a localized candidate region corresponds to a meaningful change class. In our setting, vision-language encoders are not used to replace segmentation, but to refine it. They provide the semantic screening stage of the curation pipeline, allowing us to filter noisy candidates before applying retrieval-augmented Best-of-$N$ validation to ambiguous cases.

\subsection{LLM Judgment and Retrieval-Augmented Validation}

Recent work has explored using LLMs as judges rather than only as generators. In this setting, the model is conditioned on labeled examples and asked to assess a new candidate according to the same criteria. For example, \citep{lee2024vlmjudge} studies LLM-based evaluation with user-defined scoring rules and example judgments. More broadly, few-shot multimodal prompting has been shown to improve consistency and alignment in model-based judgment. Formally, let $\mathcal{E} = \{(x_k, y_k, s_k)\}_{k=1}^K$ denote a set of example image-text-score triples, and let $(x_q, y_q)$ be a query pair, such as before/after crops from a candidate change region. A judge model $J_\phi$ predicts a score 
\begin{equation}
  s_q = J_\phi((x_q, y_q)\mid \mathcal{E})  
\end{equation}
The examples in $\mathcal{E}$ act as in-context references, encouraging the judge to apply a consistent scoring criterion to the query. To further improve consistency, recent works combine judging with retrieval-augmented context~\citep{kazoom2025dontlagrag,kazoom2025vault}. Instead of scoring a query in isolation, the model is provided with similar retrieved examples from a gallery or external memory. In remote sensing, RS-RAG~\citep{wen2025rsrag} constructs a multimodal retrieval database from image-text pairs, while ImageRAG~\citep{zhang2024imagerag} retrieves contextual patches to guide reasoning over large-scale scenes. In our setting, retrieval is used to ground semantic validation of localized candidate changes. Given a gallery $\mathcal{D} = \{(x_i, y_i, t_i)\}$ and a query $(x_q, y_q)$, we retrieve the top-$R$ most similar examples according to an embedding-based similarity function, and use them as context for the judge model. This retrieval-augmented formulation is especially useful for ambiguous candidate regions, where isolated scoring may be unstable.

\subsection{Best-of-N Ranking as Preference-Guided Selection}

Preference-based learning commonly studies how to favor high-quality outputs relative to a reference generator~\citep{bai2022helpful,ouyang2022training}. Let $\pi_\theta(y|x)$ denote a policy over candidate outputs $y$ given input $x$, and let $\pi_{\mathrm{ref}}(y|x)$ be a reference policy. Under a reward model $R(x,y)$ and a KL regularization term, the standard objective is to maximize the expected reward while remaining close to $\pi_{\mathrm{ref}}$. The corresponding optimal policy has the form $\pi^\star(y|x) \propto \pi_{\mathrm{ref}}(y|x)\exp(R(x,y)/\beta)$, where $\beta$ controls the regularization strength. This shows that high-reward candidates receive higher probability mass, while low-quality candidates are suppressed. Rather than optimizing a policy through iterative updates, our framework uses this idea at inference time through \textbf{Best-of-$N$ ranking}. Given an input $x$, we sample candidate outputs $\{y_1,\dots,y_N\}$ from $\pi_{\mathrm{ref}}$, score them with a retrieval-augmented reward model $R(x,y)$, and retain the highest-scoring candidate or filter candidates above a threshold. This procedure can be viewed as an efficient approximation to preference-weighted selection without parameter updates. In our benchmark construction pipeline, Best-of-$N$ ranking serves as the core mechanism for resolving ambiguous semantic changes and retaining high-confidence annotations~\citep{jinnai2025regularized,sessa2024bond,gui2024bonbon,zhang2025reinforcement}.

\section{Methodology}

Figure~\ref{fig:pipeline_overview} presents the overall curation pipeline used to construct \textit{RSRCC}, while Figure~\ref{fig:no_change_filter} illustrates representative no-change cases that are often falsely identified as semantic changes. Our framework is designed as a hierarchical semi-supervised curation pipeline centered on \textbf{Best-of-$N$ ranking}. It combines four stages: semantic segmentation for candidate localization, connected component analysis for region extraction, image-text semantic screening, and retrieval-augmented \textbf{Best-of-$N$} validation for ambiguous cases.

\paragraph{Transformer-Based Segmentation Model}
\label{subsec:transformer_segmentation}

First, we localize candidate change regions through semantic segmentation. We employ a transformer-based segmentation model with a ViT-L encoder and a lightweight ViT-Lite decoder. The encoder extracts multi-scale visual features $\phi(\mathcal{I})$ from an input image $\mathcal{I}$, and the decoder maps them to dense semantic predictions $S \in \mathbb{R}^{H \times W \times K}$ through upsampling and feature fusion. The predicted per-pixel semantic distribution is given by $\hat{y}_{i,j}^{(k)} = \mathrm{softmax}(\psi(\phi(\mathcal{I})))$, where $\psi(\cdot)$ denotes the segmentation head and $k \in \{1,\ldots,K\}$ indexes the semantic classes. To address class imbalance in remote sensing imagery, we train the model with Dice loss, $\mathcal{L}_{\mathrm{Dice}} = 1 - \frac{1}{K}\sum_{k=1}^{K} \frac{2 \sum_{i=1}^{H}\sum_{j=1}^{W} y_{i,j}^{(k)} \hat{y}_{i,j}^{(k)}}{\sum_{i=1}^{H}\sum_{j=1}^{W} y_{i,j}^{(k)} + \sum_{i=1}^{H}\sum_{j=1}^{W} \hat{y}_{i,j}^{(k)} + \epsilon}$, where $\epsilon$ is a constant for numerical stability.

\paragraph{Object-Level Change Detection}
\label{subsec:connected_components}

After obtaining semantic segmentation masks for each image pair, we extract candidate change regions using connected component analysis~\citep{rosenfeld1966sequential}. This step converts pixel-level mask differences into coherent object-level proposals, which serve as candidate semantic changes for dataset curation. Algorithm~\ref{alg:components} summarizes the filtering procedure, which suppresses spurious noise and retains structurally meaningful regions. Formally, let $M_t, M_{t+\Delta t} \in \{0,1,\ldots,K\}^{H \times W}$ denote the predicted semantic masks at times $t$ and $t+\Delta t$. We define the difference mask as $D = \mathbb{I}[M_t \neq M_{t+\Delta t}]$, where $D \in \{0,1\}^{H \times W}$ marks pixels whose semantic labels differ across time. Connected component partitions $D$ into disjoint regions $\mathcal{C} = \{C_1, C_2, \ldots, C_N\}$, where each $C_i$ denotes a contiguous candidate change region. Each region is then filtered using adaptive thresholds $\tau_{\text{min}}$, $\tau_{\text{iou}}$, and $\tau_{\text{changed}}$, which remove small regions, enforce spatial consistency, and ensure that the region contains a sufficient proportion of changed pixels.

\begin{algorithm}[H]
\caption{Connected Components Filtering}
\label{alg:components}
\begin{algorithmic}[1]
\STATE \textbf{Input:} Difference mask $D$, segmentation masks $M_1, M_2$, label set $\mathcal{L}$
\STATE \textbf{Output:} Valid change instances $\mathcal{C}$
\STATE Initialize $\mathcal{C} \leftarrow \emptyset$
\FOR{each class $k \in \mathcal{L}$}
    \STATE Connected components $\{C_i\}$ in $\{M_1 = k\} \lor \{M_2 = k\}$
    \FOR{each $C_i$}
        \STATE Compute region size $|C_i|$; \textbf{if} $|C_i| < \tau_{\text{min}}^{A(k)}$ \textbf{continue}
        \STATE Compute overlap $\mathrm{IoU}(C_i)=\frac{|M_1 \cap M_2|}{|M_1 \cup M_2|}$ and changed-pixel ratio $p_{\text{changed}}=\sum(D \cap C_i)$
        \STATE \textbf{if} $\mathrm{IoU}(C_i) < \tau_{\text{iou}}^{A(k)}$ \textbf{or} $p_{\text{changed}} < \tau_{\text{changed}}^{A(k)}$ \textbf{continue}
        \STATE Add $C_i$ to $\mathcal{C}$
    \ENDFOR
\ENDFOR
\STATE \textbf{Return} $\mathcal{C}$
\end{algorithmic}
\end{algorithm}

\paragraph{Semantic Filtering with Image-Text Encoder}
\label{subsec:semantic_filtering}

To refine the candidate regions, we apply an image-text semantic filtering stage using a SigLIP-based vision-language encoder~\citep{zhai2023siglip} fine-tuned on remote sensing imagery~\citep{barzilai2025recipe}. Let $f_{\theta}(x)$ and $g_{\theta}(t)$ denote the image and text embedding functions for an image patch $x$ and a class prompt $t$, respectively. Given cropped patches $(x_{\text{before}}, x_{\text{after}})$, we encode the visual inputs with $f_{\theta}$ and compare them to the text embeddings of the class prompts $\{t_1, t_2, \ldots, t_n\}$ using cosine similarity. If the expected class for the observed change does not appear among the top-$k$ predictions of $x_{\text{after}}$ with similarity above a threshold $\tau_{\text{enc}}$, the candidate is discarded as a likely false change. Otherwise, it is retained for further validation. This stage provides a fast semantic screening step before the more expensive RAG Best-of-$N$ validation.

\begin{algorithm}[H]
\caption{End-to-End Dataset Creation Pipeline}
\label{alg:dataset_creation}
\begin{algorithmic}[1]
\STATE \textbf{Input:} Thresholds $\{\tau_{\text{min}}^{A(k)}, \tau_{\text{iou}}, \tau_{\text{changed}}^{A(k)}\}$
\STATE \textbf{Output:} Labeled change set $\mathcal{D}$
\STATE $\mathcal{D} \leftarrow \emptyset$
\FOR{each image pair}
    \STATE Compute $M_t$, $M_{t+\Delta t}$, and $D=\mathbb{I}[M_t \neq M_{t+\Delta t}]$
    \STATE Extract candidate regions $\mathcal{C}$ using Algorithm~\ref{alg:components}
    \FOR{each $C_i \in \mathcal{C}$}
        \STATE $(x_b,x_a) \leftarrow \textsc{Crop}(C_i)$
        \IF{$t_{\text{orig}} \notin \text{Top-}k(f_{\text{enc}}(x_b,x_a))$}
            \STATE $\mathcal{D} \leftarrow \mathcal{D} \cup \{(x_b,x_a)\}$
        \ELSIF{$J_\phi(x_b,x_a \mid \textsc{Retrieve}(t_{\text{orig}})) > \tau$}
            \STATE $\mathcal{D} \leftarrow \mathcal{D} \cup \{(x_b,x_a)\}$
        \ENDIF
    \ENDFOR
\ENDFOR
\STATE \textbf{Return} $\mathcal{D}$
\end{algorithmic}
\end{algorithm}

\paragraph{Best-of-$N$ Ranking with Large Language Models}
\label{subsec:rltr}

A second validation stage is invoked when the image-text encoder yields \emph{ambiguous} semantic evidence. Ambiguity arises when the same class prompt $t$ appears among the top-$k$ predictions for both the before and after crops, so that encoder-level similarity alone cannot reliably determine whether the region corresponds to a true semantic change or to a temporally stable region with similar appearance. Let $q = (x_{\text{before}}, x_{\text{after}}, t)$
denote such an ambiguous query, where $x_{\text{before}}$ and $x_{\text{after}}$ are the localized image crops and $t$ is the hypothesized semantic class. To reduce decision variance in this regime, we perform retrieval-augmented \textbf{Best-of-$N$ ranking} using a LLM. Given the query $q$, a retrieval module selects a context set $\mathcal{E}_R(q)=\{(x_i, y_i, s_i)\}_{i=1}^{R}$ of semantically similar annotated examples from a gallery, where each tuple contains a reference example, its associated label or explanation, and an optional supervision signal. The retrieved set acts as an in-context support set that conditions the judge on examples lying near the local semantic neighborhood of the query. The judge then defines a scalar relevance or consistency score $r(q) = J_{\phi}(q \mid \mathcal{E}_R(q)),$ where larger values indicate stronger evidence that the candidate corresponds to a valid change of class $t$. If multiple competing candidate hypotheses
\begin{equation}
\mathcal{H}(q)=\{h_1,\dots,h_N\}
\end{equation}
are generated for the same ambiguous region, with each $h_j$ representing a distinct interpretation or candidate label, we score each hypothesis as
\begin{equation}
r_j = J_{\phi}(h_j \mid \mathcal{E}_R(q)),
\qquad j=1,\dots,N,
\end{equation}
and apply the Best-of-$N$ decision rule $ h^\star = \arg\max_{h_j \in \mathcal{H}(q)} r_j.$ More generally, when the objective is confidence-based filtering rather than single-hypothesis selection, we retain only those candidates satisfying $r_j > \tau,$ for a fixed threshold $\tau$. This procedure can be interpreted as an inference-time surrogate for preference-guided selection: instead of updating model parameters, we rank or filter candidate semantic interpretations according to a retrieval-conditioned reward signal. Retrieval is essential here because it constrains the judgment to locally relevant reference cases, thereby reducing instability that may arise when ambiguous regions are scored in isolation. In our pipeline, this Best-of-$N$ stage serves as the main ambiguity-resolution mechanism and substantially improves the semantic precision of the final curated benchmark. Prompt details are provided in Appendix~\ref{subsec:prompt_engineering}, and additional implementation details appear in Appendix~\ref{subsec:custom_dataset_new}.

\paragraph{Question-Answer Extraction}
\label{subsec:qa_extraction}

After identifying candidate change regions $\mathcal{B} = \{b_1, b_2, \ldots, b_N\}$, we use a LLM to generate question-answer pairs for each localized change instance. Each region $b_i$ is overlaid on the corresponding image pair $(x_{\text{before}}, x_{\text{after}})$ so that the model receives a region-grounded view of the detected change. Given the cropped visual input $\mathcal{I}_i = \{x_{\text{before}}^{(b_i)}, x_{\text{after}}^{(b_i)}\}$ and the pipeline-derived class label $c_i$, the model generates a question-answer pair $(q_i, a_i)$. We generate:

\noindent\textbf{Closed-ended (MCQ):} 4-option questions offering discrete semantic interpretations, where $q_i \in \mathcal{Q}_{\text{mc}}$.

\noindent\textbf{Yes/No:} binary questions about the presence or absence of a semantic change, where $q_i \in \mathcal{Q}_{\text{yn}}$.

\noindent\textbf{Open-ended:} free-form descriptive questions for linguistic diversity, where $q_i \in \mathcal{Q}_{\text{open}}$.

Importantly, the LLM is not used to determine whether a change occurred. The ground-truth semantic label $c_i$ is derived solely from the vision-based curation pipeline. The LLM serves only as a linguistic diversification module, generating natural and varied phrasings conditioned on the localized region and its validated semantic label. Formally, $(q_i, a_i) = f_{\text{LLM}}(\mathcal{I}_i, c_i)$, where $f_{\text{LLM}}$ denotes the question-generation function. It therefore evaluates whether a model can reason about a particular semantic change instance rather than only summarize global scene differences. Additional qualitative examples are provided in Appendix~\ref{sec:qualitative_examples}.

\section{Results}

\noindent\textbf{Dataset.}
The \textit{LEVIR-CD} dataset consists of high-resolution satellite image pairs ($512 \times 512$ px, $\approx 0.5$\,\text{m/pixel}) capturing diverse urban and suburban scenes. 
Although LEVIR-CD follows a fixed resolution, our framework is resolution-agnostic: all filtering, similarity, and region-based operations are normalized by image dimensions, enabling deployment across datasets with varying spatial scales.

\noindent\textbf{Models.}
For the retrieval-augmented LLM validation we use Gemma-3-4B~\citep{gemma2024}.
For captioning we use Gemini-2.5-Flash~\citep{comanici2025gemini25}.

\noindent We evaluate our framework both quantitatively and qualitatively to assess scalability, reliability, and semantic consistency using the following metrics:

\noindent\textit{Human Agreement (\%)} - percentage of evaluators agreeing with generated questions or answers.

\noindent\textit{Accuracy (\%)} - correctness of binary (Yes/No) and multiple-choice responses.

\noindent\textit{BLEU} - $n$-gram precision measuring lexical overlap between model-generated answers and the ground-truth answers in the dataset.

\noindent\textit{BERTScore (F1)} - semantic similarity between model-generated and human-written captions.

\noindent\textit{CIDEr} - consensus-based metric that measures similarity between generated captions and multiple human references using TF-IDF weighted $n$-grams.

\noindent\textit{SPICE} - semantic propositional metric that evaluates agreement between generated and reference captions at the level of objects, attributes, and relations.

\begin{table}[H]
\centering
\scriptsize
\caption{
Human evaluation across dataset creation pipelines.
\textbf{Ours (Segmentation + Encoder + Best-of-$N$ Ranking)} achieves the highest agreement.
}
\label{tab:creation_pipeline}
\begin{tabular}{p{2cm}|p{4cm}|p{4cm}}
\toprule
\textbf{Creation Pipeline} & \textbf{Human-Agreement (Zero-Shot)}~($\uparrow$) & \textbf{Human-Agreement (Few-Shot)}~($\uparrow$) \\
\midrule
\textbf{Ours} & \multicolumn{2}{c}{\textbf{98\%}} \\
\midrule
Encoder Only & \multicolumn{2}{c}{\textbf{11\%}} \\
\midrule
Segmentation Only & \multicolumn{2}{c}{\textbf{61\%}} \\
\midrule
Gemma-3-4B & 27\% & 31\% \\
Gemma-3-27B & 29\% & 37\% \\
Gemini-2.5-Flash & 51\% & 57\% \\
Gemini-2.5-Pro & 55\% & 59\% \\
\bottomrule
\end{tabular}
\end{table}

\begin{table}[H]
\centering
\scriptsize

\caption{Evaluation of human and model performance. Accuracy is shown for binary tasks and BERTScore for open-ended responses.}

\label{tab:overall_results}
\begin{tabular}{p{1.7cm}p{1.9cm}p{1cm}p{2cm}p{2.2cm}}
\toprule
\textbf{Question Type} & \textbf{Metric} & \textbf{Change}~($\uparrow$) & \textbf{No Change}~($\uparrow$) & \textbf{Total Accuracy}~($\uparrow$) \\
\midrule
Yes/No          & Accuracy  & 91.33\% & 93.33\% & 92.33\% \\
Multiple-Choice & Accuracy  & 95.33\% & 99.00\% & 97.17\% \\
Open-Ended      & BERTScore & 76.89\% & 98.12\% & 87.51\% \\
\bottomrule
\end{tabular}
\end{table}

\noindent\textbf{Table~\ref{tab:creation_pipeline}} presents human evaluation results for different stages of the dataset creation pipeline, serving as a direct measure of change detection capability. Each isolates one component of the pipeline to measure its independent contribution to overall agreement. The \textit{few-shot} configuration denotes that the LLM received example pairs with corresponding questions before evaluation, improving its contextual understanding, while ours achieves the highest score, confirming the complementary role of each stage in producing reliable change annotations. Our method significantly outperforms all baselines across the board. \textbf{Table~\ref{tab:overall_results}} reports human agreement results measuring annotator acceptance of the generated questions and answers. High scores across binary, multiple-choice, and open-ended formats indicate that human evaluators consistently found the outputs coherent, relevant, and aligned with the visual content. These results validate the linguistic quality and factual consistency of the generated dataset, demonstrating strong human-model agreement across all task types.

\begin{table*}[!ht]
\centering
\tiny
\caption{Baseline benchmark performance across LLMs and question types. Metrics include Accuracy, BERTScore, and BLEU-$n$ (1–4) for change and no-change cases, with total accuracy summarizing overall performance.}
\label{tab:benchmark_results}
\begin{tabular}{p{2cm}|p{2.3cm}p{1.7cm}p{1.7cm}p{1.7cm}p{2cm}}
\toprule
\textbf{LLM Used} & \textbf{Question Type} & \textbf{Metric} & \textbf{Change}~($\uparrow$) & \textbf{No Change}~($\uparrow$) & \textbf{Total}~($\uparrow$) \\
\midrule

\multirow{9}{*}{\textbf{Gemini-2.5-Flash}}
& Yes/No Questions & Accuracy & 67.02 & 38.96 & 65.65 \\
& Multiple-Choice Questions & Accuracy & 46.93 & 28.71 & 46.04 \\
& Open-Ended Questions & BERTScore & 35.97 & 32.28 & 35.77 \\
& Open-Ended Questions & BLEU-1 & 88.37 & 77.37 & 87.93 \\
& Open-Ended Questions & BLEU-2 & 73.39 & 66.79 & 72.96 \\
& Open-Ended Questions & BLEU-3 & 32.21 & 21.83 & 32.00 \\
& Open-Ended Questions & BLEU-4 & 12.74 & 15.61 & 12.83 \\
& Open-Ended Questions & CIDEr & 58.73 & 55.94 & 57.18 \\
& Open-Ended Questions & SPICE & 21.37 & 20.42 & 21.05 \\

\midrule

\multirow{9}{*}{\textbf{Gemini-2.5-Pro}}
& Yes/No Questions & Accuracy & 69.82 & 39.11 & 68.32 \\
& Multiple-Choice Questions & Accuracy & 49.73 & 29.85 & 48.76 \\
& Open-Ended Questions & BERTScore & 37.77 & 33.89 & 37.57 \\
& Open-Ended Questions & BLEU-1 & 92.79 & 81.24 & 92.30 \\
& Open-Ended Questions & BLEU-2 & 77.06 & 70.13 & 76.64 \\
& Open-Ended Questions & BLEU-3 & 33.82 & 22.93 & 33.60 \\
& Open-Ended Questions & BLEU-4 & 13.37 & 16.39 & 13.47 \\
& Open-Ended Questions & CIDEr & 67.82 & 64.55 & 66.19 \\
& Open-Ended Questions & SPICE & 24.91 & 23.76 & 24.34 \\

\midrule

\multirow{9}{*}{\textbf{Gemma-3-4B}}
& Yes/No Questions & Accuracy & 47.72 & 29.21 & 46.81 \\
& Multiple-Choice Questions & Accuracy & 38.92 & 19.75 & 37.98 \\
& Open-Ended Questions & BERTScore & 29.58 & 26.21 & 29.41 \\
& Open-Ended Questions & BLEU-1 & 71.23 & 68.00 & 71.08 \\
& Open-Ended Questions & BLEU-2 & 58.46 & 61.28 & 58.60 \\
& Open-Ended Questions & BLEU-3 & 21.89 & 19.87 & 21.79 \\
& Open-Ended Questions & BLEU-4 & 07.58 & 08.67 & 07.63 \\
& Open-Ended Questions & CIDEr & 46.85 & 44.27 & 45.56 \\
& Open-Ended Questions & SPICE & 18.73 & 17.42 & 18.06 \\

\midrule

\multirow{9}{*}{\textbf{Gemma-3-27B}}
& Yes/No Questions & Accuracy & 49.72 & 31.97 & 48.85 \\
& Multiple-Choice Questions & Accuracy & 31.25 & 23.73 & 30.88 \\
& Open-Ended Questions & BERTScore & 31.28 & 28.07 & 31.12 \\
& Open-Ended Questions & BLEU-1 & 76.85 & 67.28 & 76.38 \\
& Open-Ended Questions & BLEU-2 & 63.82 & 58.08 & 63.55 \\
& Open-Ended Questions & BLEU-3 & 28.01 & 18.99 & 27.57 \\
& Open-Ended Questions & BLEU-4 & 11.08 & 13.58 & 11.21 \\
& Open-Ended Questions & CIDEr & 46.85 & 44.27 & 45.56 \\
& Open-Ended Questions & SPICE & 18.73 & 17.42 & 18.06 \\

\bottomrule
\end{tabular}
\end{table*}

\textbf{Table~\ref{tab:benchmark_results}} presents baseline results on our test benchmark, reporting off-the-shelf model performance across multiple question types and evaluation metrics. In addition to Accuracy, BLEU-$n$, and BERTScore, we report CIDEr and SPICE to better capture consensus with human annotations and semantic consistency. These results establish reference scores for different LLMs.

\section{Conclusions and Future Work}

\noindent
We presented \textit{RSRCC}, a benchmark for remote sensing change question-answering constructed through a hierarchical semi-supervised curation pipeline centered on Best-of-$N$ ranking. By combining segmentation-based candidate localization, vision-language semantic filtering, and retrieval-augmented Best-of-$N$ validation, our framework produces reliable and semantically consistent annotations at scale. The resulting dataset provides a benchmark resource for evaluating multimodal models on localized, change-specific reasoning, rather than only global image-level change description. Our current framework remains tied to a predefined set of semantic classes, and may underperform on novel, visually ambiguous, or compositional changes outside this label space. In future work, we plan to expand the benchmark with additional classes, explore semi-supervised and open-vocabulary annotation strategies, and evaluate broader cross-model generalization across multiple LLMs and vision-language encoders. More broadly, we hope \textit{RSRCC} will help advance remote sensing change captioning from coarse scene-level summaries toward fine-grained, region-grounded semantic understanding.

\clearpage

\bibliographystyle{plainnat}
\bibliography{neurips_2026}

\clearpage

\appendix

\section{Group-Restricted Retrieval and Boundary Preservation}
\label{sec:group_restricted_retrieval}

We formalize the intuition that conditioning on retrieved examples drawn from the model's \emph{own} knowledge groups reduces bias by avoiding out-of-distribution (OOD) exemplars.
Intuitively, the encoder partitions the data manifold into representation groups (clusters) that share semantics and appearance.
Retrieval constrained to the query's group preserves local decision geometry, whereas conditioning on examples from other groups can shift the effective decision boundary.

\paragraph{Setup.}
Let $\mathcal{X}$ be the input space of image pairs and let $h:\mathcal{X}\rightarrow\mathbb{R}^d$ be the fixed vision encoder.
Let $\mathcal{D}$ denote the in-distribution data manifold.
Assume $\mathcal{D}$ admits a (possibly unknown) partition into $G$ groups
\begin{equation}
\mathcal{D}=\bigcup_{g=1}^{G}\mathcal{D}_g,\qquad \mathcal{D}_g\cap\mathcal{D}_{g'}=\emptyset \ \text{for}\ g\neq g',
\end{equation}
where each $\mathcal{D}_g$ corresponds to a coherent semantic/visual mode (a ``knowledge group'').
For a query $x\in\mathcal{D}$, let $g(x)$ be its (latent) group index, i.e., $x\in \mathcal{D}_{g(x)}$.

We consider a downstream predictor $F(\cdot;\mathcal{S})$ (e.g., the multimodal reasoning module) that produces an output for $x$ conditioned on a set $\mathcal{S}$ of retrieved exemplars.
Let $\ell(\cdot,\cdot)$ be a bounded loss, and define the conditional risk
\begin{equation}
\mathcal{R}(x;\mathcal{S})=\mathbb{E}\big[\ell(F(x;\mathcal{S}),y)\mid x\big].
\end{equation}

\paragraph{Group-restricted retrieval.}
Given a similarity kernel in embedding space, e.g.,
\begin{equation}
k(x,r)=\exp\!\left(-\frac{\|h(x)-h(r)\|^2}{\tau}\right),
\end{equation}
standard retrieval selects $\mathcal{S}$ from the $K$ nearest neighbors of $x$ under $\|h(x)-h(r)\|$.
We define the \emph{group-restricted} retrieval set
\begin{equation}
\mathcal{N}_K^{\mathrm{in}}(x)=\Big\{r\in\mathcal{D}_{g(x)}:\ r \ \text{is among the $K$ nearest neighbors of $x$}\Big\},
\end{equation}
and the complementary \emph{cross-group} set
\begin{equation}
\mathcal{N}_K^{\mathrm{out}}(x)=\Big\{r\in\mathcal{D}\setminus\mathcal{D}_{g(x)}:\ r \ \text{is among the $K$ nearest neighbors of $x$}\Big\}.
\end{equation}

\paragraph{Boundary shift under cross-group conditioning.}
We assume the conditional predictor is Lipschitz in a summary statistic of retrieved exemplars.
Let $\phi(\mathcal{S})\in\mathbb{R}^m$ be a representation of the retrieved context (e.g., pooled retrieved embeddings or retrieved captions), and assume
\begin{equation}
\|F(x;\mathcal{S})-F(x;\mathcal{S}')\|\leq L_F \cdot \|\phi(\mathcal{S})-\phi(\mathcal{S}')\|
\end{equation}
for all valid $\mathcal{S},\mathcal{S}'$.
Define the within-group diameter in embedding space
\begin{equation}
\Delta_{\mathrm{in}}(x)=\max_{r\in \mathcal{D}_{g(x)}} \|h(x)-h(r)\|,
\end{equation}
and the group separation margin
\begin{equation}
\Delta_{\mathrm{out}}(x)=\min_{r\in \mathcal{D}\setminus \mathcal{D}_{g(x)}} \|h(x)-h(r)\|.
\end{equation}
When the encoder induces well-separated groups, we have $\Delta_{\mathrm{out}}(x)\gg \Delta_{\mathrm{in}}(x)$ for typical $x$.

\textbf{Group-Restricted Retrieval Reduces Context Shift}
\label{lem:group_restricted}
Assume $\phi(\mathcal{S})$ is an average of retrieved embeddings, i.e.,
$\phi(\mathcal{S})=\frac{1}{|\mathcal{S}|}\sum_{r\in\mathcal{S}} h(r)$.
Let $\mathcal{S}_{\mathrm{in}}\subseteq \mathcal{N}_K^{\mathrm{in}}(x)$ and let $\mathcal{S}_{\mathrm{mix}}$ be any set of the same size that contains at least one cross-group exemplar, i.e., $\mathcal{S}_{\mathrm{mix}}\cap \mathcal{N}_K^{\mathrm{out}}(x)\neq \emptyset$.
Then,
\begin{equation}
\|\phi(\mathcal{S}_{\mathrm{in}})-h(x)\|\leq \Delta_{\mathrm{in}}(x),
\end{equation}
whereas
\begin{equation}
\|\phi(\mathcal{S}_{\mathrm{mix}})-h(x)\|\geq \frac{1}{|\mathcal{S}_{\mathrm{mix}}|}\Big(\Delta_{\mathrm{out}}(x)-(|\mathcal{S}_{\mathrm{mix}}|-1)\Delta_{\mathrm{in}}(x)\Big).
\end{equation}
Consequently, when $\Delta_{\mathrm{out}}(x)$ dominates $\Delta_{\mathrm{in}}(x)$, cross-group conditioning induces a larger context shift in the encoder space.

\paragraph{Implication for bias and robustness.}
By the Lipschitz property of $F$,
Lemma~\ref{lem:group_restricted} yields a direct bound on prediction shift:
\begin{equation}
\|F(x;\mathcal{S}_{\mathrm{in}})-F(x;\mathcal{S}_{\mathrm{mix}})\|
\leq L_F \cdot \|\phi(\mathcal{S}_{\mathrm{in}})-\phi(\mathcal{S}_{\mathrm{mix}})\|.
\end{equation}
Thus, restricting retrieval to $\mathcal{D}_{g(x)}$ controls the induced context shift and helps preserve the local decision boundary around $x$.
In contrast, conditioning on cross-group examples may move the effective context away from $h(x)$, increasing boundary distortion and introducing bias due to OOD exemplars.

In our framework, retrieval is performed in the encoder space and the Best-of-$N$ filter further selects candidates whose retrieved context yields semantically consistent change descriptions.
Together, these mechanisms promote in-group conditioning and reduce the likelihood of spurious reasoning driven by out-of-group examples.

\section{Inference-Time Preference-Guided Ranking}
\label{sec:inference_policy_improvement}

Our framework employs an inference-time preference-guided ranking strategy to select high-quality captions without performing gradient-based optimization.
Let $\pi_{\text{ref}}(y|s)$ denote a frozen generator that produces candidate captions $y \in \mathcal{Y}$ for an image pair $s \in \mathcal{S}$.
Each candidate is evaluated using a retrieval-augmented scoring function $J_{\phi}(s,y)$, which measures semantic and visual consistency conditioned on reference examples $\mathcal{E}$.

\subsection{Preference-Weighted Scoring}

Following standard preference learning formulations~\citep{ouyang2022training}, we consider the regularized objective
\begin{equation}
\max_{\pi} \;
\mathbb{E}_{s \sim \mathcal{D},\, y \sim \pi(\cdot|s)}
\left[
J_{\phi}(s,y)
- \beta \log \frac{\pi(y|s)}{\pi_{\text{ref}}(y|s)}
\right],
\label{eq:pref_objective}
\end{equation}
where $\beta$ controls the strength of regularization.

\subsubsection{Closed-Form Preference Distribution}

As shown in prior work~\citep{bai2022helpful}, the optimal solution to Eq.~\ref{eq:pref_objective} admits the closed-form expression
\begin{equation}
\pi^{\star}(y|s)
=
\frac{1}{Z(s)}\,
\pi_{\text{ref}}(y|s)\,
\exp\!\left(\frac{J_{\phi}(s,y)}{\beta}\right),
\label{eq:closed_form}
\end{equation}
where $Z(s)$ is a normalization constant.

This formulation corresponds to a preference-weighted re-ranking of candidate captions, increasing the probability of semantically consistent hypotheses.

\subsection{Best-of-N Approximation}

Direct sampling from $\pi^{\star}$ is computationally expensive.
Instead, we approximate Eq.~\ref{eq:closed_form} via Best-of-$N$ sampling.
For each input $s$, we sample
\begin{equation}
\{y_1,\ldots,y_N\} \sim \pi_{\text{ref}}(\cdot|s),
\end{equation}
and compute corresponding scores
\begin{equation}
r_i = J_{\phi}(s,y_i), \quad i=1,\ldots,N.
\end{equation}

The final prediction is obtained through thresholding or maximum selection:
\begin{equation}
y^{\star} = \arg\max_{y_i} r_i, 
\qquad \text{or} \qquad r_i > \tau.
\label{eq:bon_selection}
\end{equation}

\begin{figure}[!t]
\centering
\includegraphics[width=0.95\linewidth]{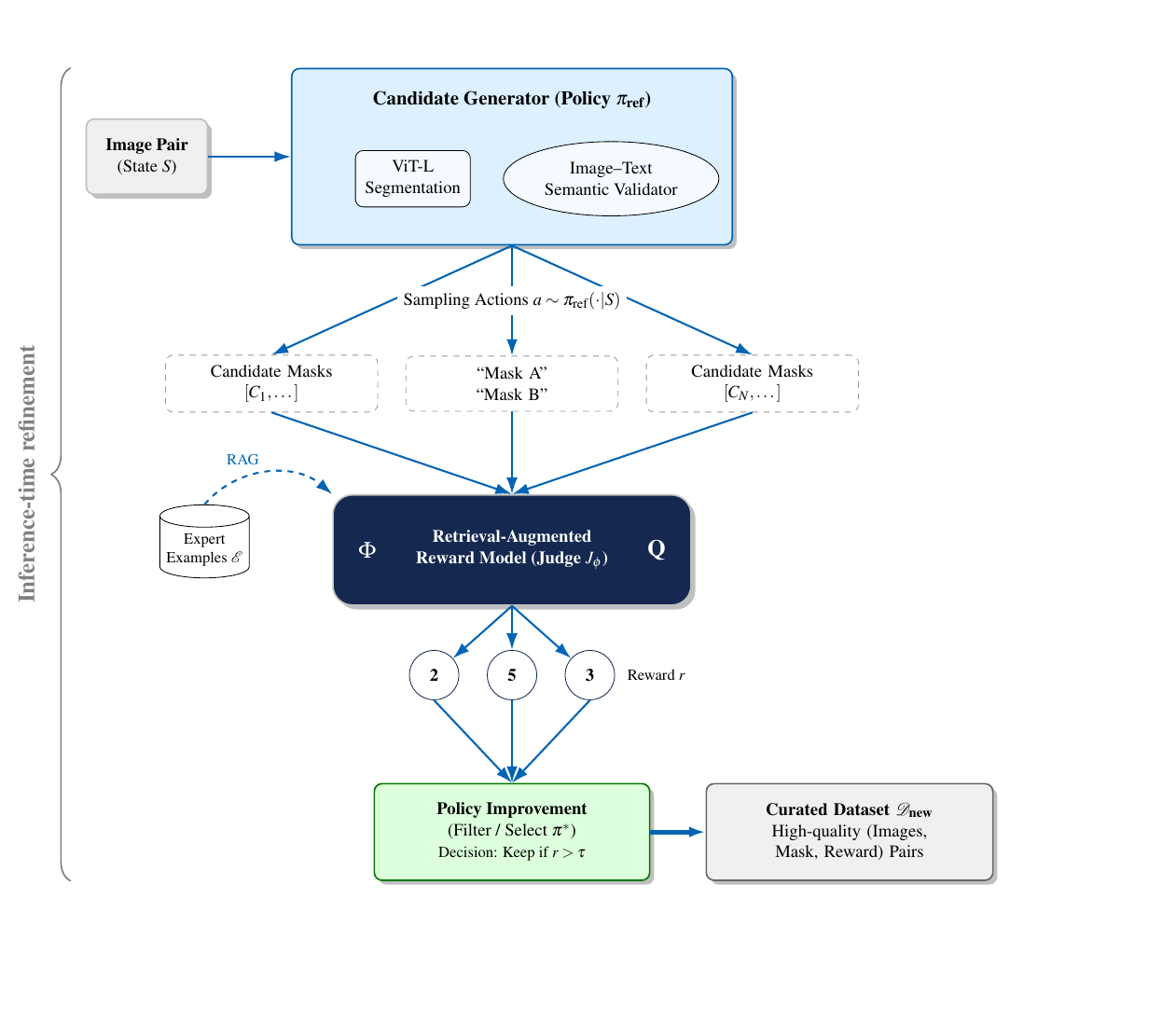}
\caption{Illustration of the preference-guided Best-of-$N$ selection process.
A frozen generator produces multiple candidate captions, which are scored using a retrieval-augmented preference model.
High-scoring candidates are retained to form the refined dataset.}
\label{fig:rl_refinement}
\end{figure}

This procedure corresponds to rejection sampling from the preference-weighted distribution, enabling efficient approximation of $\pi^{\star}$ without parameter updates.

\subsection{Dataset Refinement}

The curated dataset is formed by retaining high-scoring samples
\begin{equation}
\mathcal{D}_{\text{new}} = \{(s,y_i)\mid r_i>\tau\},
\end{equation}
which concentrates probability mass on semantically reliable annotations.
This inference-time refinement improves dataset quality while preserving stability and reproducibility.

Figure~\ref{fig:rl_refinement} illustrates the overall preference-guided selection process.

\section{Theoretical Interpretation of Best-of-$N$ Filtering}
\label{sec:theory_bestofn}

We provide a simplified interpretation of why the proposed filtering mechanism remains stable under segmentation and localization noise.

Let $x$ denote an input image pair, and let $\mathcal{C}(x)$ be a set of candidate regions generated by a segmentation model.
Due to imperfect localization, $\mathcal{C}(x)$ may contain both relevant and irrelevant regions.

Let $f(c)$ denote the score assigned to a candidate region $c$ by the downstream retrieval and reasoning module, measuring its semantic consistency with the observed change.
We assume that for true change regions $c^\ast$, $f(c^\ast)$ is stochastically larger than for irrelevant regions.

The Best-of-$N$ filtering procedure selects
\begin{equation}
c^\star = \arg\max_{c \in \mathcal{C}_N(x)} f(c),
\end{equation}
where $\mathcal{C}_N(x)$ denotes a randomly sampled subset of $N$ candidates.

Under mild assumptions on the score distribution, the probability that $c^\star$ corresponds to a true change region increases exponentially with $N$.
Consequently, Best-of-$N$ acts as a noise suppression mechanism that amplifies semantically consistent regions while discarding spurious candidates.

This interpretation explains the observed robustness to segmentation backbone choice and random cropping, as reported in Section~\ref{sec:random_crops}. Without Best-of-$N$, we observe frequent failure cases in which visually salient but semantically irrelevant regions dominate the reasoning process.
This confirms that the filtering stage is necessary rather than merely beneficial.

\section{Qualitative Comparison: Missed Changes}
\label{app:no_change_comparison}

\begin{figure}[!ht]
    \centering
    \includegraphics[width=0.8\linewidth]{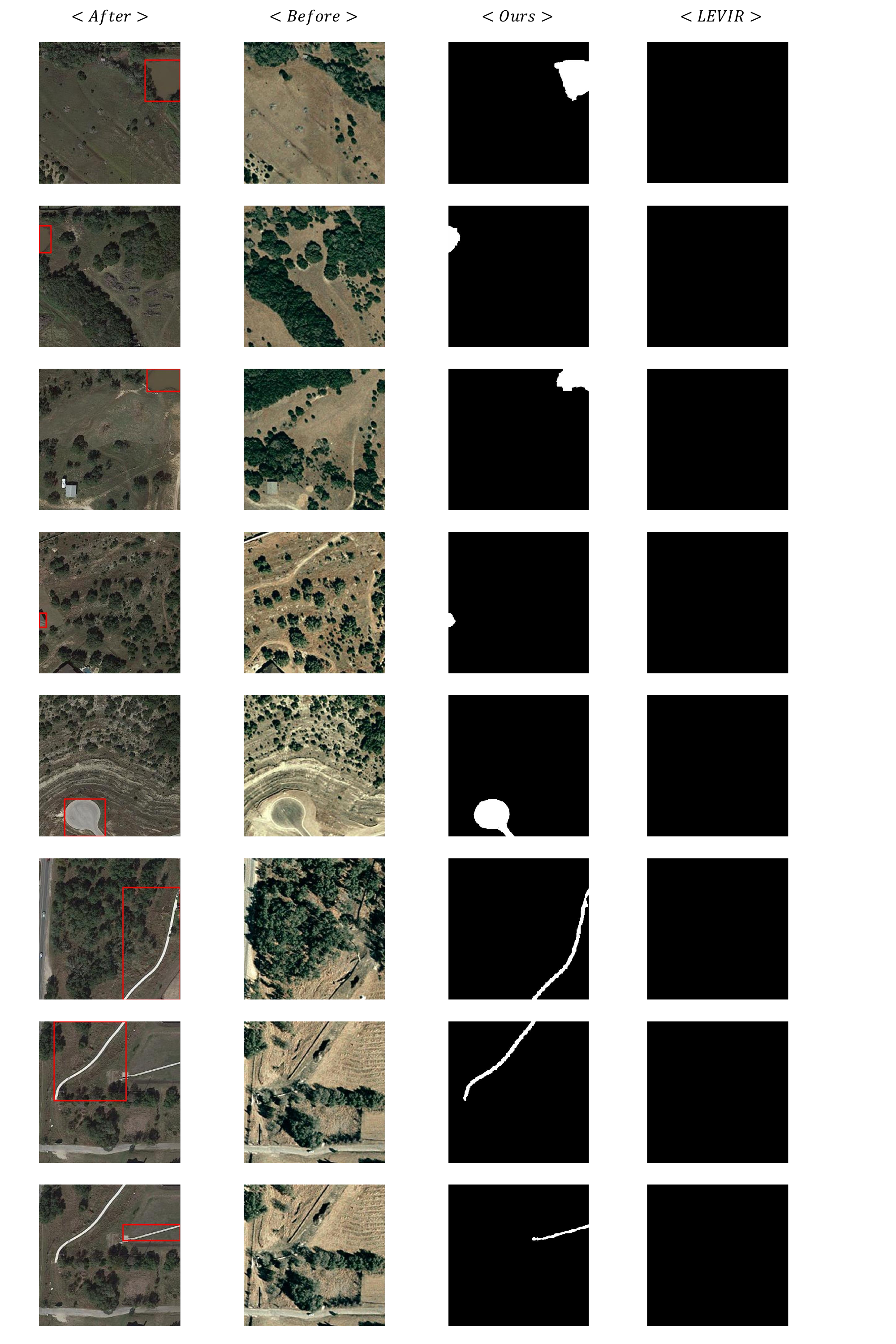}
    \caption{Qualitative comparison between our method and the \textsc{LEVIR-CC} ground truth.  
    Our model detects subtle structural or environmental changes that are annotated as ``no change'' in the public dataset.}
    \label{fig:no_change_gt_vs_ours}
\end{figure}

Figure~\ref{fig:no_change_gt_vs_ours} illustrates qualitative examples highlighting cases where our method successfully detects meaningful semantic changes that are labeled as ``no change'' in the public \citep{chen2020levircd} dataset.  
While others provide strong building-level annotations, it occasionally omits fine-grained structural or surface-level changes (e.g., new roads, terrain modifications, or partial constructions).  

Our segmentation pipeline demonstrates greater sensitivity to these overlooked regions, producing accurate masks that capture finer structural or environmental variations.  
Although we do not report a direct metric comparison against the ground truth, we qualitatively observe that our method yields more semantically consistent results, especially in cases where \textsc{LEVIR-CC} annotations fail to reflect real-world changes.

\subsection{Detection of Small-Scale Changes}
\label{subsec:veget_detection_comparison}

\begin{figure}[!ht]
    \centering
    \includegraphics[width=0.8\linewidth]{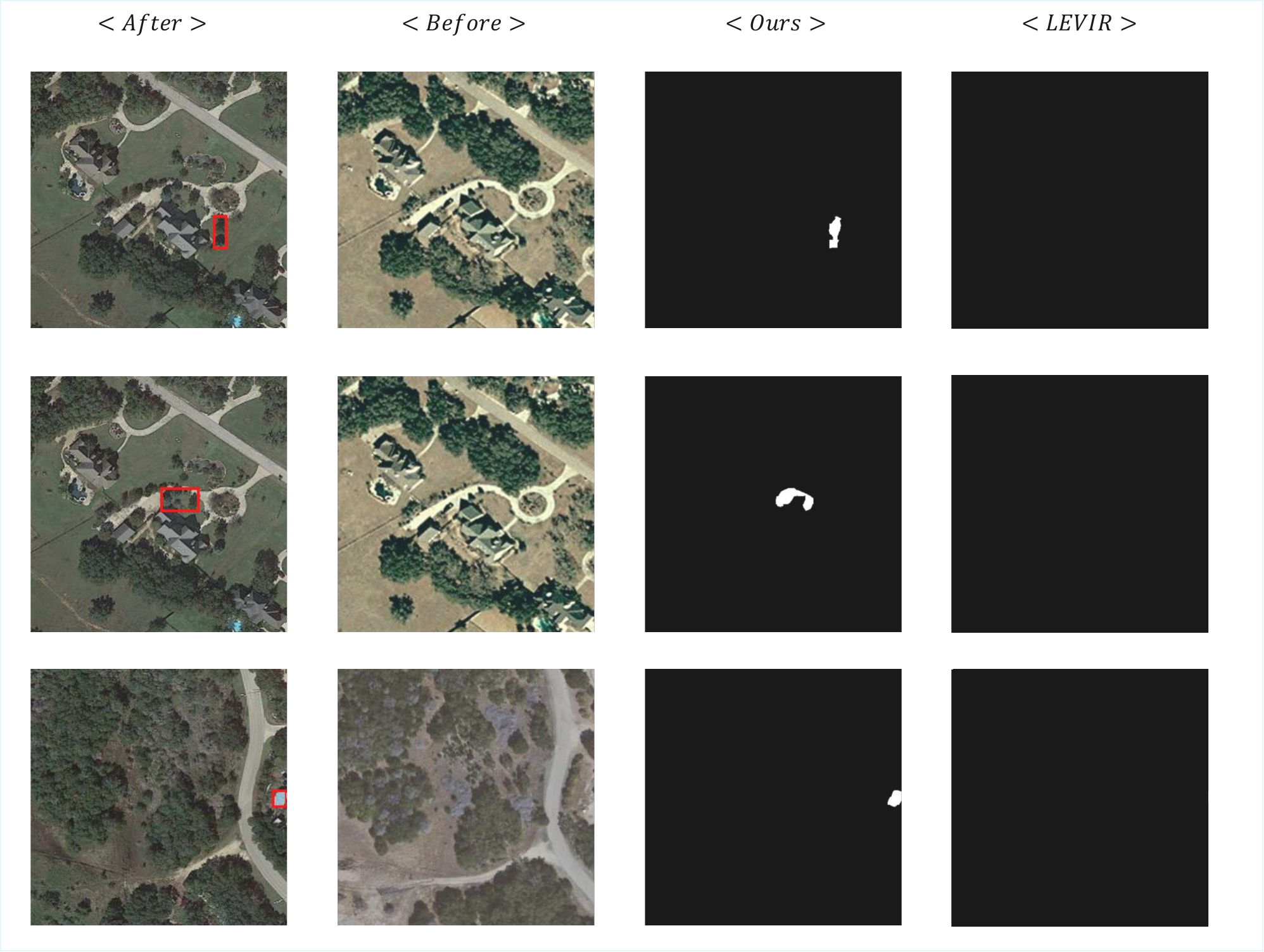}
    \caption{
    Comparison of small-scale change detection in vegetation and buildings.
    Examples~1 and~2 show successful detection of vegetation changes,
    while example~3 captures a subtle building modification.
    Our method detects these fine-grained changes that are commonly overlooked
    in \textit{LEVIR} due to incomplete ground truth annotations.
    }
    \label{fig:veget_detection_comparison}
\end{figure}

\noindent
To assess the robustness of our method in identifying fine-grained semantic changes, 
we evaluate its ability to detect subtle variations in vegetation, building structures, and newly constructed roads. 
As shown in Figure~\ref{fig:veget_detection_comparison}, 
examples~1 and~2 highlight vegetation and tree-level changes that our method successfully detects but are missing in the \textit{LEVIR} annotations, 
while example~3 demonstrates a minor building modification accurately captured by our pipeline.  
These examples underscore the advantage of our segmentation- and vision–language–guided framework, 
which remains sensitive to localized and small-scale variations often misrepresented in existing ground truth datasets.  

\begin{figure}[!ht]
    \centering
    \includegraphics[width=0.8\linewidth]{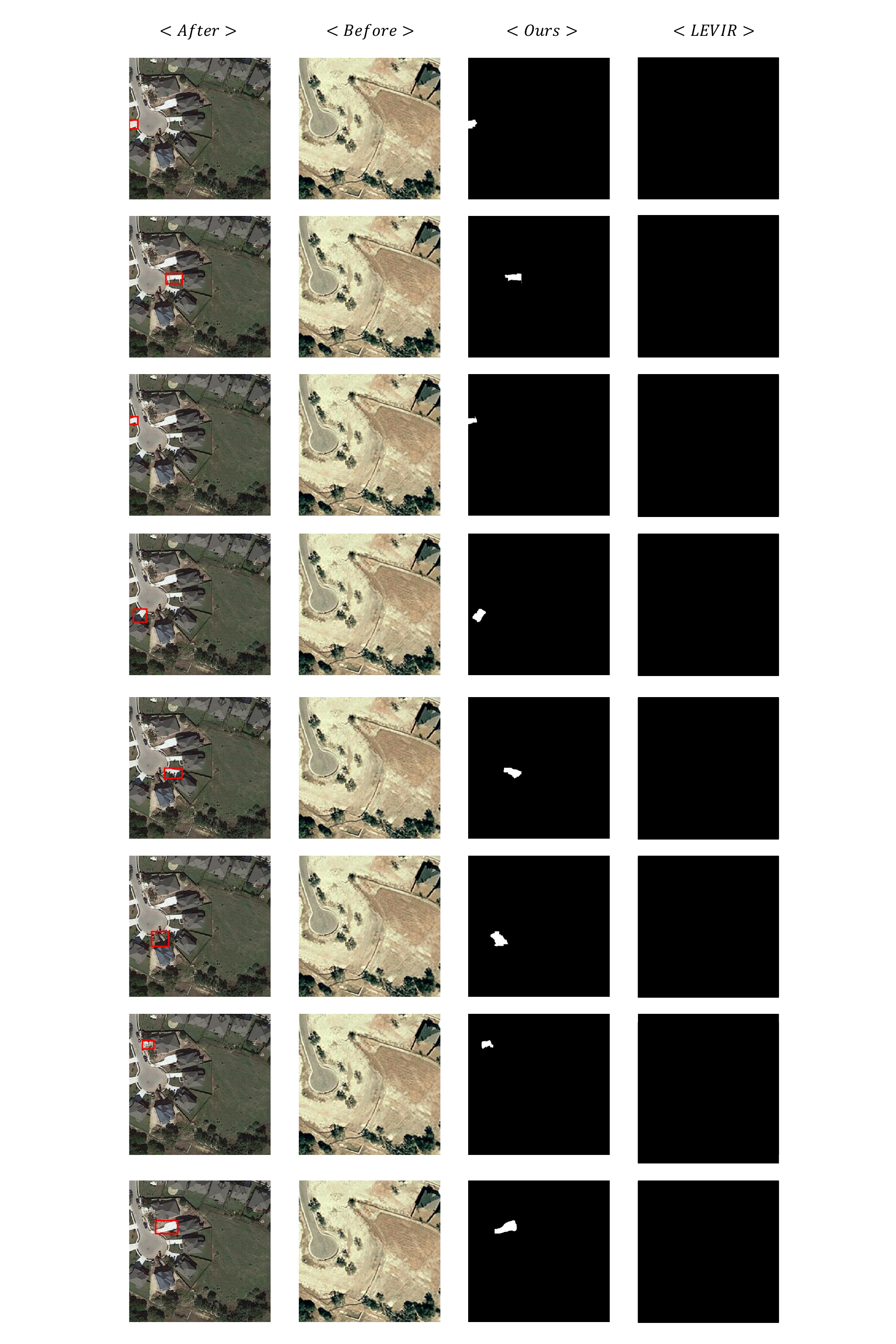}
    \caption{
    Detection of newly constructed or extended roads.  
    Our approach correctly identifies subtle road-level changes that prior methods missed, primarily due to incomplete or misaligned ground truth annotations in \textit{LEVIR}. 
    }
    \label{fig:roads_detection}
\end{figure}

Furthermore, Figure~\ref{fig:roads_detection} presents cases of newly built or extended roads that our approach identifies correctly, 
while previous methods classify them as \textit{no-change} due to incomplete ground truth coverage.  
This emphasizes our model’s capability to generalize to fine-grained, low-contrast structural updates, even in challenging rural or semi-urban regions.

\section{Human Agreement Evaluation of Preference-Filtered Pseudo-Labels}
\label{subsec:filter_analysis}

\subsection{On Convergence under Noisy Pseudo-Labels}
\label{subsubsec:convergence_lemma}

We formalize the intuition that preference-guided filtering yields high-quality pseudo-labels whose aggregate supervision remains statistically aligned with true semantic changes, even when individual annotations are imperfect.

\textbf{Convergence under High-Confidence Pseudo-Labels}
Let $(X,Y)$ be random variables where $X \in \mathcal{X}$ denotes an input region and $Y \in \{0,1\}$ indicates the true change label.
Let $\tilde{Y} \in \{0,1\}$ denote the pseudo-label produced by our segmentation and ranking pipeline, and let $D \in \{0,1\}$ indicate whether a sample is retained after preference-based filtering.

Assume
\begin{equation}
\mathbb{P}(\tilde{Y}=Y \mid X, D=1) \geq 1 - \varepsilon,
\quad \forall X \in \mathcal{X},
\end{equation}
for some $\varepsilon < \frac{1}{2}$, and that retained samples
$\{(X_i,\tilde{Y}_i): D_i=1\}_{i=1}^{n}$
are drawn i.i.d.\ from $\mathbb{P}(X,\tilde{Y}\mid D=1)$.

Let $\mathcal{H}$ be a hypothesis class with finite capacity and let
\begin{equation}
h_n \in \arg\min_{h \in \mathcal{H}}
\frac{1}{n}\sum_{i=1}^{n}\ell(h(X_i),\tilde{Y}_i)
\end{equation}
denote the empirical risk minimizer.
Then $h_n$ converges in probability to a Bayes-optimal classifier.

\noindent
Under standard learning assumptions, training on sufficiently many retained pseudo-labels yields a consistent estimator, despite occasional labeling noise.

\medskip
\noindent\textit{Informal Interpretation.}
Preference-based filtering ensures that retained pseudo-labels are biased toward correct semantic interpretations.
When learning from a large collection of such high-confidence but imperfect samples, the aggregated supervision remains aligned with the true decision boundary.
Consequently, the model learns dominant visual patterns of semantic change rather than overfitting spurious artifacts.

\subsection{Human Agreement Protocol}

To assess pseudo-label quality, we construct a held-out evaluation subset.
For each candidate region $s_i$ produced by the pipeline, a human annotator provides a binary judgment indicating whether the region corresponds to a meaningful and visually coherent semantic change.
Since dense pixel-level annotations are unavailable, evaluation is conducted at the region level.

\medskip
\noindent\textbf{Binary Agreement Labels.}
Each evaluated region is assigned
\begin{equation}
a_i \in \{0,1\},
\end{equation}
where $a_i=1$ indicates human agreement.
The filtering procedure produces a retention indicator
\begin{equation}
d_i \in \{0,1\},
\end{equation}
where $d_i=1$ denotes that the region is retained.

\medskip
\noindent\textbf{Agreement Precision.}
We measure the fraction of retained regions judged as valid:
\begin{equation}
P_{\mathrm{HA}} =
\frac{\sum_i d_i a_i}{\sum_i d_i}.
\end{equation}

A higher $P_{\mathrm{HA}}$ indicates that the preference-guided filtering mechanism effectively prioritizes semantically meaningful change regions over visually ambiguous or trivial segments.

\subsubsection{Effect of Best-of-$N$ Sampling}

We provide a simple theoretical justification for the effectiveness of Best-of-$N$ sampling in improving the reliability of retained candidates.

Let $r(y)$ denote the preference score assigned to a candidate region or caption $y$, and let $\tau$ be a fixed acceptance threshold.
Assume that candidates $\{y_i\}_{i=1}^{N}$ are drawn independently from a proposal distribution $\pi_{\text{ref}}(\cdot|s)$ conditioned on an input $s$.

\textbf{Best-of-$N$ Improves Acceptance Probability}
\label{lem:bon_accept}
Let
\begin{equation}
p_{\tau} = \mathbb{P}\big(r(y) > \tau\big)
\end{equation}
denote the probability that a single sample exceeds the acceptance threshold.
Then, the probability that at least one out of $N$ samples is accepted satisfies
\begin{equation}
\mathbb{P}\Big(\max_{1\le i\le N} r(y_i) > \tau\Big)
=
1 - (1 - p_{\tau})^{N}.
\end{equation}
In particular, this probability is monotonically increasing in $N$.

\noindent
The result follows directly from the independence assumption and shows that increasing the number of sampled candidates exponentially reduces the probability of rejecting all valid hypotheses.
Consequently, Best-of-$N$ sampling improves the likelihood of retaining high-quality semantic change annotations without modifying the underlying generator.

\section{Satellite Image Acquisition and Preprocessing}
\label{subsec:custom_dataset_new}
We constructed a custom dataset to exclude regions dominated by vegetation or trees in both temporal images 
(before and after change). To achieve this, we applied a patch filtering procedure based on statistical and 
colorimetric thresholds, as detailed in Algorithm~\ref{alg:filtering}. This filtering ensures that only 
urban or built-up regions with minimal vegetation are retained for further analysis.

For comparison and benchmarking, we additionally employed the LEVIR-CD dataset~\citep{chen2020levircd}, 
a widely used benchmark for building change detection in very high-resolution (VHR) remote sensing imagery. 
LEVIR-CD consists of paired satellite images captured at different times over the same geographic regions, 
with pixel-level annotations of building changes. This dataset enables the study of semantic changes over 
time under realistic urban conditions. 

Importantly, we present all LEVIR-CD samples \textit{as is}, without applying any of our filtering criteria, 
to maintain consistency with prior work and preserve the original dataset distribution.

\begin{enumerate}
    \item \textbf{Uniformity filter:} Patches with low pixel intensity variation are discarded. If
    % \[
    \begin{equation}
    \sigma(X) < \tau_{\text{std}}, \quad \tau_{\text{std}} = 0.08,
    \end{equation}
    % \]
    the patch is excluded.

    \item \textbf{Low saturation filter:} Patches with low mean saturation in HSV space are removed. If
    % \[
    \begin{equation}
    \frac{1}{HW} \sum_{i=1}^H \sum_{j=1}^W S_{ij} < \tau_{\text{sat}}, \quad \tau_{\text{sat}} = 0.15,
    \end{equation}
    % \]
    the patch is discarded.

    \item \textbf{Vegetation filter:} Using the Excess Green Index,
    % \[
    \begin{equation}
    \text{ExG} = 2G - R - B,
    \end{equation}
    % \]
    patches dominated by vegetation are removed if
    % \[
    \begin{equation}
    \frac{1}{HW} \sum_{i=1}^H \sum_{j=1}^W \text{ExG}_{ij} > \tau_{\text{ExG}}, \quad \tau_{\text{ExG}} = 0.35.
    \end{equation}
    % \]
\end{enumerate}

\begin{algorithm}[H]
\caption{Patch Filtering Process}
\label{alg:filtering}
\begin{algorithmic}[1]
\STATE \textbf{Input:} Image patch $X \in \mathbb{R}^{H \times W \times 3}$, thresholds $\tau_{\text{std}}, \tau_{\text{sat}}, \tau_{\text{ExG}}$
\STATE \textbf{Output:} Keep or discard decision
\STATE Compute pixel intensity standard deviation $\sigma(X)$
\IF{$\sigma(X) < \tau_{\text{std}}$}
    \STATE Discard patch
\ENDIF
\STATE Convert $X$ to HSV, compute mean saturation $\bar{S}$
\IF{$\bar{S} < \tau_{\text{sat}}$}
    \STATE Discard patch
\ENDIF
\STATE Compute Excess Green Index: $\text{ExG} = 2G - R - B$
\STATE Compute mean $\overline{\text{ExG}}$
\IF{$\overline{\text{ExG}} > \tau_{\text{ExG}}$}
    \STATE Discard patch
\ENDIF
\STATE Keep patch
\end{algorithmic}
\end{algorithm}

\subsection{IoU Threshold Calibration}

To determine an appropriate intersection-over-union (IoU) threshold for identifying true object-level changes, 
we conducted a controlled calibration experiment using human evaluation over a randomly selected subset of 
100 image pairs. Each pair was manually annotated to indicate whether the observed differences corresponded 
to genuine structural change (e.g., new construction, building removal) or false change 
(e.g., illumination shifts, vegetation growth, or shadow variations). 
We then compared these binary human judgments against a continuous range of IoU values 
computed between masks at times $t$ and $t + \Delta t$.

\begin{figure}[!ht]
    \centering
    \includegraphics[width=0.6\textwidth]{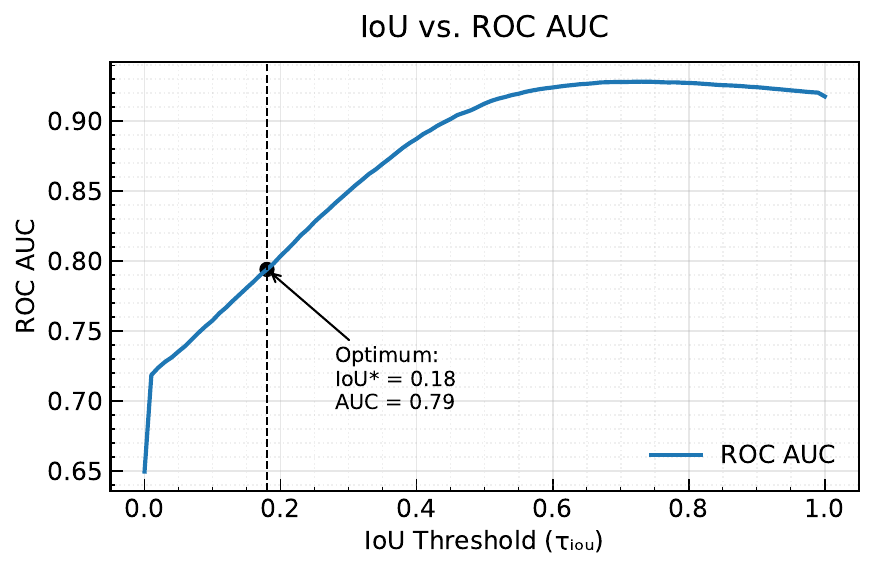}
    \caption{
    \textbf{IoU vs. ROC AUC.} ROC-AUC analysis over the IoU range $[0, 1]$ using human-annotated subtest images.
    The peak AUC is achieved near $\tau_{\text{iou}} = 0.18$, indicating this value yields the best trade-off 
    between sensitivity and specificity in detecting true structural changes.
    }
    \label{fig:iou_auc_curve}
\end{figure}

Figure~\ref{fig:iou_auc_curve} illustrates the receiver operating characteristic (ROC) analysis 
used to optimize the IoU threshold $\tau_{\text{iou}}$. The ROC curve evaluates the trade-off between 
true positive rate (TPR) and false positive rate (FPR) when varying $\tau_{\text{iou}}$ over the interval 
$[0, 1]$. The analysis achieved an area under the curve (AUC) of 0.91, confirming that IoU is a strong 
discriminator of true versus false changes. The optimal threshold $\tau_{\text{iou}} = 0.18$ corresponds to 
the maximal Youden's $J$ index, representing the most balanced separation between changed and unchanged regions 
as perceived by human evaluators.

\begin{figure}[!ht]
    \centering
    \includegraphics[width=0.6\linewidth]{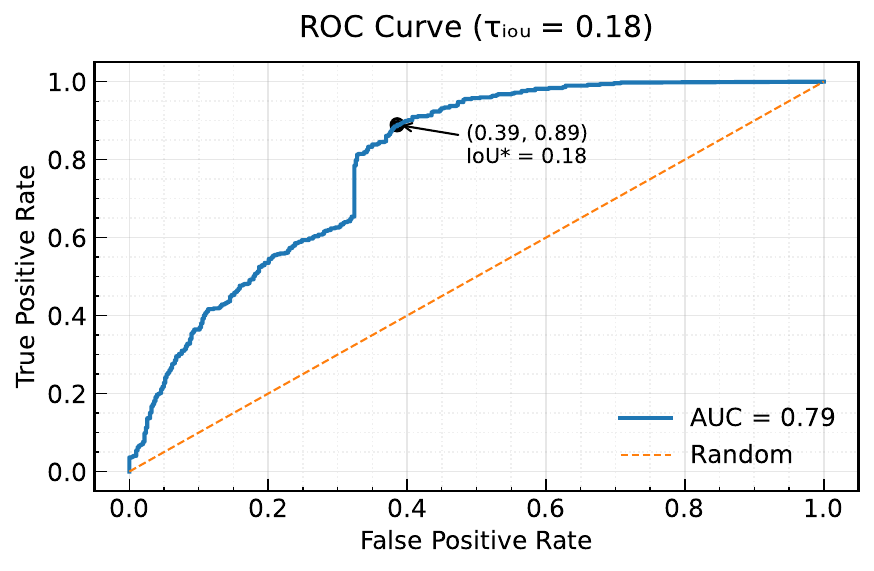}
    \caption{
    \textbf{ROC Curve for IoU Threshold Optimization.}
    The optimal intersection-over-union threshold $\tau_{\text{iou}} = 0.18$ achieves the best trade-off 
    between true and false detections, as determined from 100 human-annotated image pairs. 
    High TPR and low FPR values confirm that this threshold reliably distinguishes meaningful structural changes 
    from spurious differences.
    }
    \label{fig:roc_curve_iou}
\end{figure}

To further illustrate the decision boundary, Figure~\ref{fig:roc_curve_iou} shows the ROC curve 
corresponding to the optimal threshold $\tau_{\text{iou}} = 0.18$. 
The near-perfect separation ($\text{AUC} = 0.79$) demonstrates that human judgments align well 
with this empirical threshold, providing an interpretable and reproducible boundary for localizing 
true changes while suppressing noisy detections.

\subsection{Top-$k$ Similarity Evaluation}

To assess the reliability of the retrieval stage, we conducted a human evaluation experiment using a subset of 
100 test images. For each query patch $x$, the SigLIP model retrieved the top-$k$ candidates 
$\{x_1, x_2, \ldots, x_k\}$ ranked by cosine similarity:
\begin{equation}
s_i = \frac{f_{\theta}(x) \cdot f_{\theta}(x_i)}{\|f_{\theta}(x)\|\|f_{\theta}(x_i)\|}, \quad s_i \in [0,1].
\end{equation}
A human annotator was then asked to decide whether the retrieved patch $x_i$ corresponds to the same semantic 
class as the query image. For each value of $k \in \{1,2,\dots,10\}$, we computed the average percentage of 
retrieved samples judged as correct:
\begin{equation}
A(k) = \frac{1}{N} \sum_{n=1}^{N} \mathbbm{1}\big[\text{HumanAgree}(x_n, x_{n,i\le k})\big],
\end{equation}
where $\mathbbm{1}[\cdot]$ is an indicator function and $N=100$ is the number of evaluated images.

\begin{figure}[!ht]
    \centering
    \includegraphics[width=0.6\linewidth]{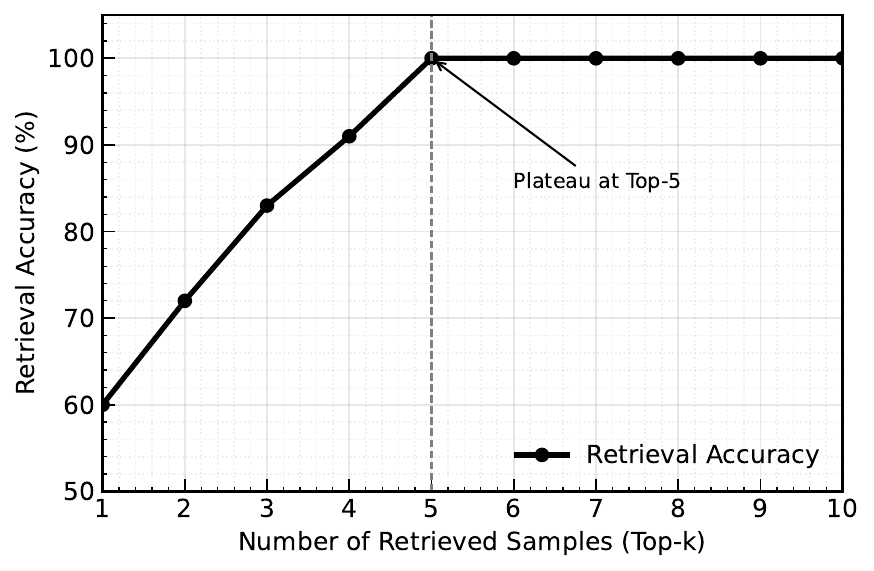}
    \caption{
    \textbf{Top-$k$ Retrieval Consistency.}
    Human-annotated retrieval accuracy for SigLIP embeddings over 100 test images. 
    Accuracy increases rapidly up to $k=5$, after which a plateau at 100\% agreement is observed, 
    confirming that Top-5 retrieval captures all semantically correct matches.
    }
    \label{fig:topk_consistency}
\end{figure}

As shown in Figure~\ref{fig:topk_consistency}, retrieval accuracy increases sharply from Top-1 to Top-5, 
then reaches a stable plateau. This indicates that for all 100 test samples, the correct semantic class 
was consistently present within the top five retrieved neighbors, with no improvement for larger $k$. 
Consequently, we fixed $k=5$ as the retrieval depth for all subsequent experiments, 
balancing recall, semantic diversity, and computational efficiency.

\subsection{Evaluation of Distance Metrics for Human-Annotated Similarity}

To assess the effectiveness of different distance metrics for similarity-based retrieval, we conducted a human evaluation on a subset of $N = 100$ test images.  
For each image $x_i$, annotators were asked to confirm whether the true class appeared within the Top-5 retrieved results under each metric.  
The final accuracy for each metric $m$ was computed as:

\begin{equation}
A(m) = \frac{1}{N} \sum_{i=1}^{N} \mathbbm{1}[\text{HumanAgree}(x_i, m)],
\end{equation}
where $\mathbbm{1}[\cdot]$ is an indicator function equal to 1 when the annotator confirmed the retrieval as correct.

\paragraph{Distance Metrics.}  
We compared four commonly used distance metrics for feature-space similarity evaluation:
\begin{itemize}
    \item \textbf{Cosine Similarity:}
    \begin{equation}
    S_{\text{cos}}(E_1, E_2) = \frac{E_1 \cdot E_2}{\|E_1\| \, \|E_2\|},
    \end{equation}
    which measures angular similarity between two embedding vectors.
    
    \item \textbf{L1 Distance:}
    \begin{equation}
    d_{L1}(E_1, E_2) = \sum_{j=1}^{D} |E_{1j} - E_{2j}|,
    \end{equation}
    providing a robust measure of absolute deviation across dimensions.

    \item \textbf{L2 Distance:}
    \begin{equation}
    d_{L2}(E_1, E_2) = \sqrt{\sum_{j=1}^{D} (E_{1j} - E_{2j})^2},
    \end{equation}
    which emphasizes larger differences by squaring deviations.

    \item \textbf{Wasserstein Distance:}
    \begin{equation}
    W(E_1, E_2) = \inf_{\gamma \in \Gamma(E_1, E_2)} \mathbb{E}_{(x, y) \sim \gamma}[\|x - y\|],
    \end{equation}
    representing the minimal transport cost required to align the distributions of two embedding sets.
\end{itemize}

\begin{figure}[!ht]
    \centering
    \includegraphics[width=0.8\linewidth]{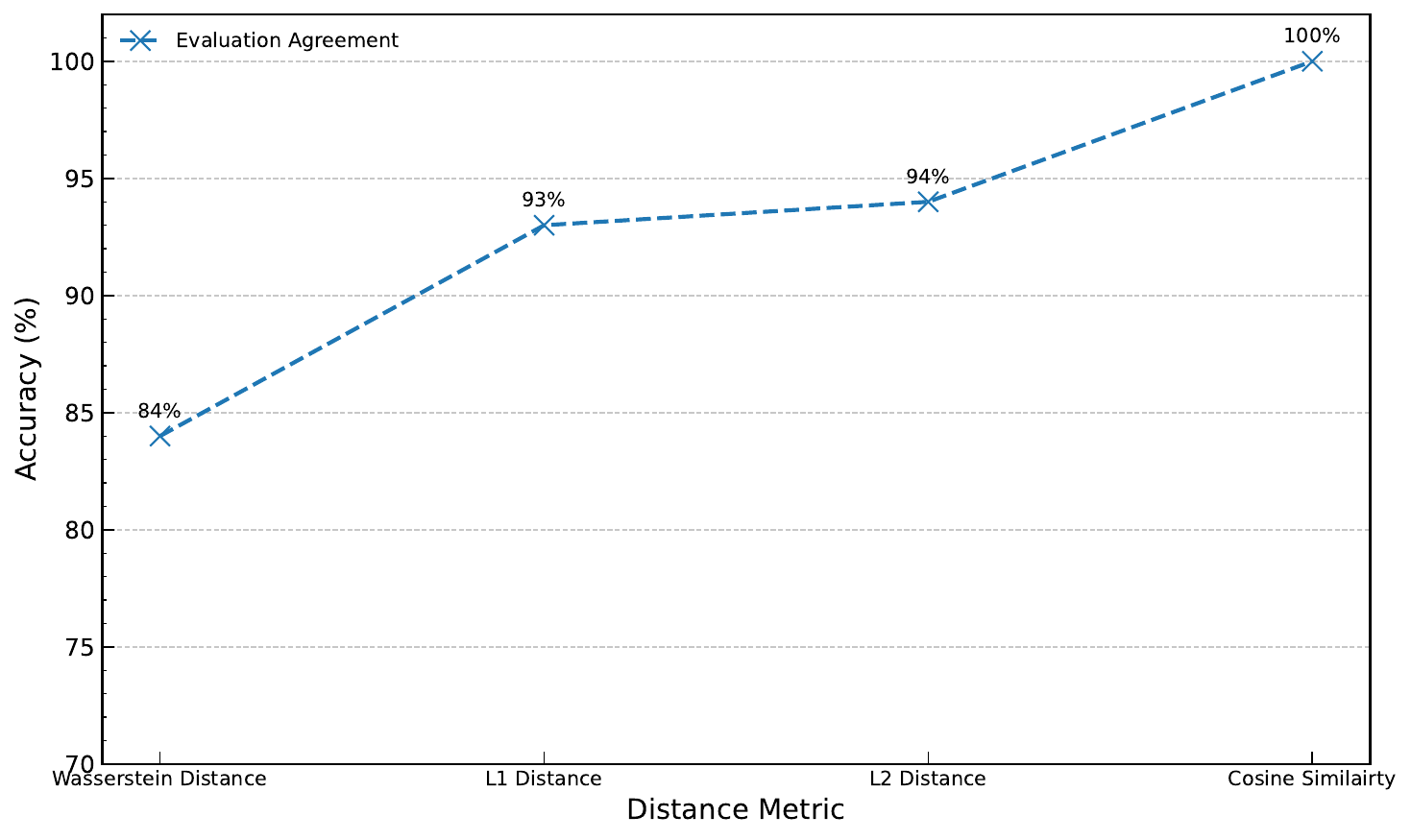}
    \caption{Human-annotated retrieval agreement across different distance metrics.  
    Accuracy values are computed for $N=100$ test images with Top-5 similarity retrieval.  
    Cosine similarity achieved the highest correspondence with human evaluation.}
    \label{fig:metric_comparison}
\end{figure}

\paragraph{Results.}  
Figure~\ref{fig:metric_comparison} illustrates the agreement rate between human annotators and each metric.  
Cosine similarity achieved perfect alignment ($100\%$ agreement), while $L_2$, $L_1$, and Wasserstein distances achieved $94\%$, $93\%$, and $84\%$ respectively.  
The results suggest that cosine similarity best captures the semantic relationships in the embedding space, while distance-based metrics are less effective in representing perceptual alignment with human judgment.

\section{Embedding Model Comparison for Top-5 Retrieval}

To evaluate the retrieval efficacy of different embedding models on our Top-$k$ similarity (Cosine Similarity) task, we compared three representative vision–language encoders: 
\textbf{CLIP}~\citep{radford2021clip}, \textbf{MaMMUT}~\citep{singh2022mammut}, and \textbf{SigLIP 2}~\citep{Tschannen2025SigLIP2}.  

CLIP jointly optimizes an image encoder $f_\theta$ and a text encoder $g_\phi$ using a contrastive loss:
\begin{equation}
\mathcal{L}_{\text{CLIP}} = - \frac{1}{N} \sum_{i=1}^{N} 
\log \frac{\exp(\text{sim}(f_\theta(I_i), g_\phi(T_i))/\tau)}
{\sum_{j=1}^{N} \exp(\text{sim}(f_\theta(I_i), g_\phi(T_j))/\tau)},
\end{equation}
where $\text{sim}$ denotes cosine similarity and $\tau$ is a temperature parameter.
SigLIP replaces the softmax loss with a sigmoid-based regression objective:
\begin{equation}
\mathcal{L}_{\text{SigLIP}} = -[y \cdot \log \sigma(s) + (1-y) \cdot \log (1-\sigma(s))],
\end{equation}
which improves training efficiency and scalability.
Finally, MaMMUT employs a multi-task loss combining masked language modeling and image–text matching:
\begin{equation}
\mathcal{L}_{\text{MaMMUT}} = \mathcal{L}_{\text{MLM}} + \mathcal{L}_{\text{MIM}} + \mathcal{L}_{\text{ITM}},
\end{equation}
enhancing fine-grained alignment for subtle semantic reasoning.

\begin{figure}[!ht]
    \centering
    \includegraphics[width=0.7\linewidth]{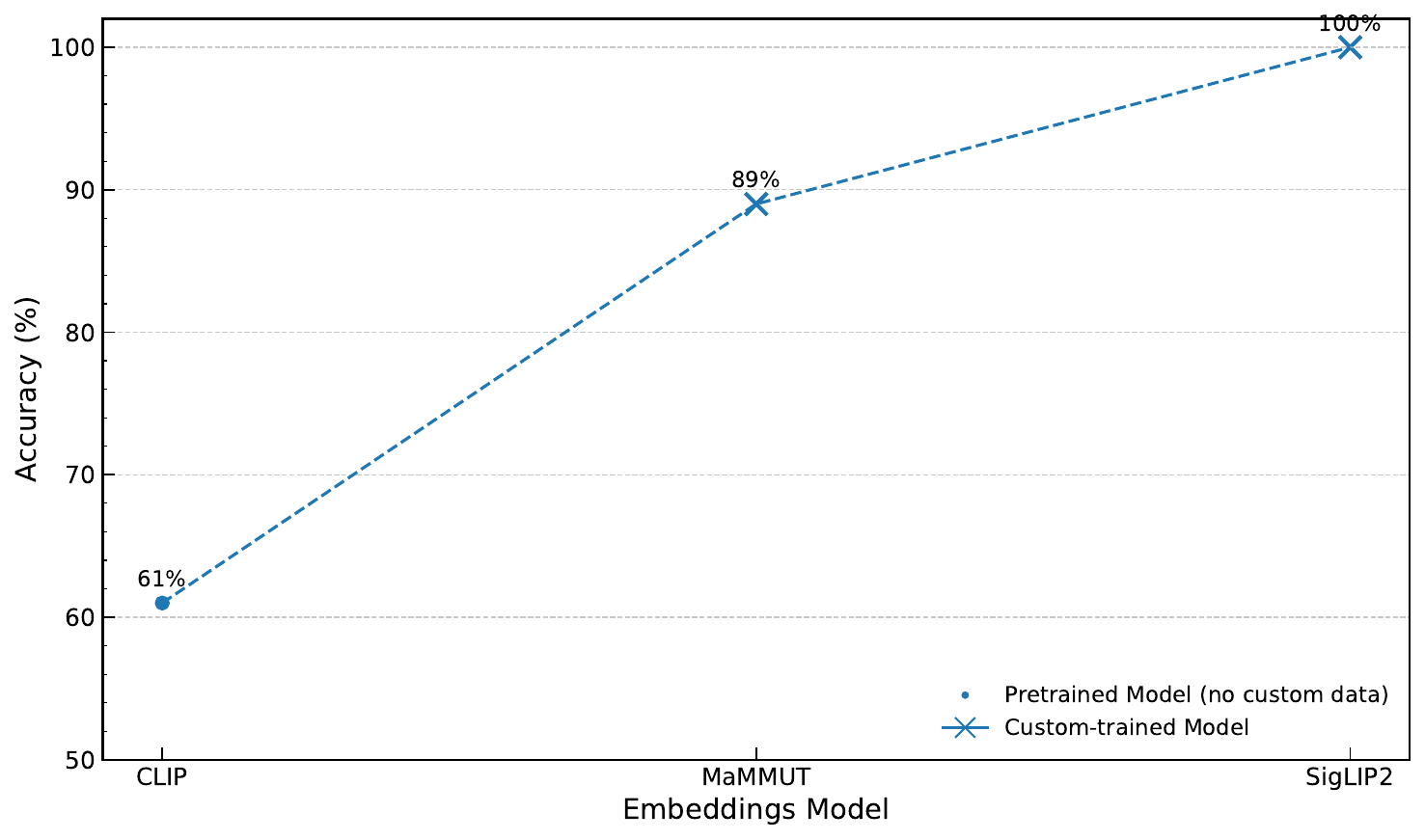}
    \caption{
    \textbf{Top-5 Retrieval Accuracy across Embedding Models.} Human annotator agreement rates for Top-5 candidate sets generated by each model.  
    Marker types differentiate: “\texttt{.}” denotes a purely pretrained model, and “\texttt{x}” denotes a model fine-tuned on our domain.
    }
    \label{fig:embeddings_comp}
\end{figure}

For each query patch, a human annotator determined whether the correct semantic class appeared within the top-5 retrieved candidates generated by each model. 
Figure~\ref{fig:embeddings_comp} presents the resulting annotation agreement rates.
As shown, CLIP achieved 61\%, MaMMUT reached 89\%, and SigLIP 2 achieved 100\%.
These results motivated our use of SigLIP 2 as the retrieval backbone for downstream ranking.

Figure~\ref{fig:pool_detection} highlights our method’s ability to adapt to small and complex change categories such as \textit{swimming pools}. 
These regions are challenging because they occupy only a few pixels relative to the full image resolution, often leading to misdetections or omissions in existing benchmarks. 
Our approach accurately segments these small changes, confirming that the adaptive filtering and hierarchical validation stages allow the model to generalize effectively even in fine-grained, low-area scenarios.

\begin{figure}[!ht]
    \centering
    \includegraphics[width=0.8\linewidth]{figs/pool.pdf}
    \caption{Visual comparison on challenging small-object cases. 
    Our method demonstrates superior adaptability in detecting fine-grained changes such as \textit{swimming pools}, which often occupy very small regions within high-resolution images. 
    Compared to \textit{LEVIR}, our approach successfully identifies subtle localized changes while avoiding false positives in surrounding areas.}
    \label{fig:pool_detection}
\end{figure}

\section{Robustness to Random Crops}
\label{sec:random_crops}

To further validate the effectiveness of the proposed filtering and retrieval pipeline, we conduct an additional robustness experiment based on randomly sampled image regions.
Specifically, we extract random spatial crops from the input images, without conditioning on any detected change regions or semantic cues, and process them through the full pipeline.

This experiment is designed to assess whether the method admits spurious or irrelevant regions as valid candidates.
Ideally, random crops that do not correspond to meaningful changes should be consistently rejected by the filtering mechanism.

Across $1{,}000$ randomly sampled crops, only $2$ samples are retained by the pipeline, corresponding to a pass rate of $0.2\%$.
This low acceptance rate indicates that the proposed Best-of-$N$ filtering strategy effectively suppresses noisy or semantically irrelevant regions, even under unconstrained random sampling.

These results further demonstrate that the method exhibits strong selectivity and robustness, reinforcing its ability to focus on meaningful change patterns while avoiding false positives induced by random spatial variations.

\section{Class Distribution Analysis}
\label{subsec:class_distribution}

\begin{figure}[!htbp]
    \centering
    \includegraphics[width=0.7\linewidth]{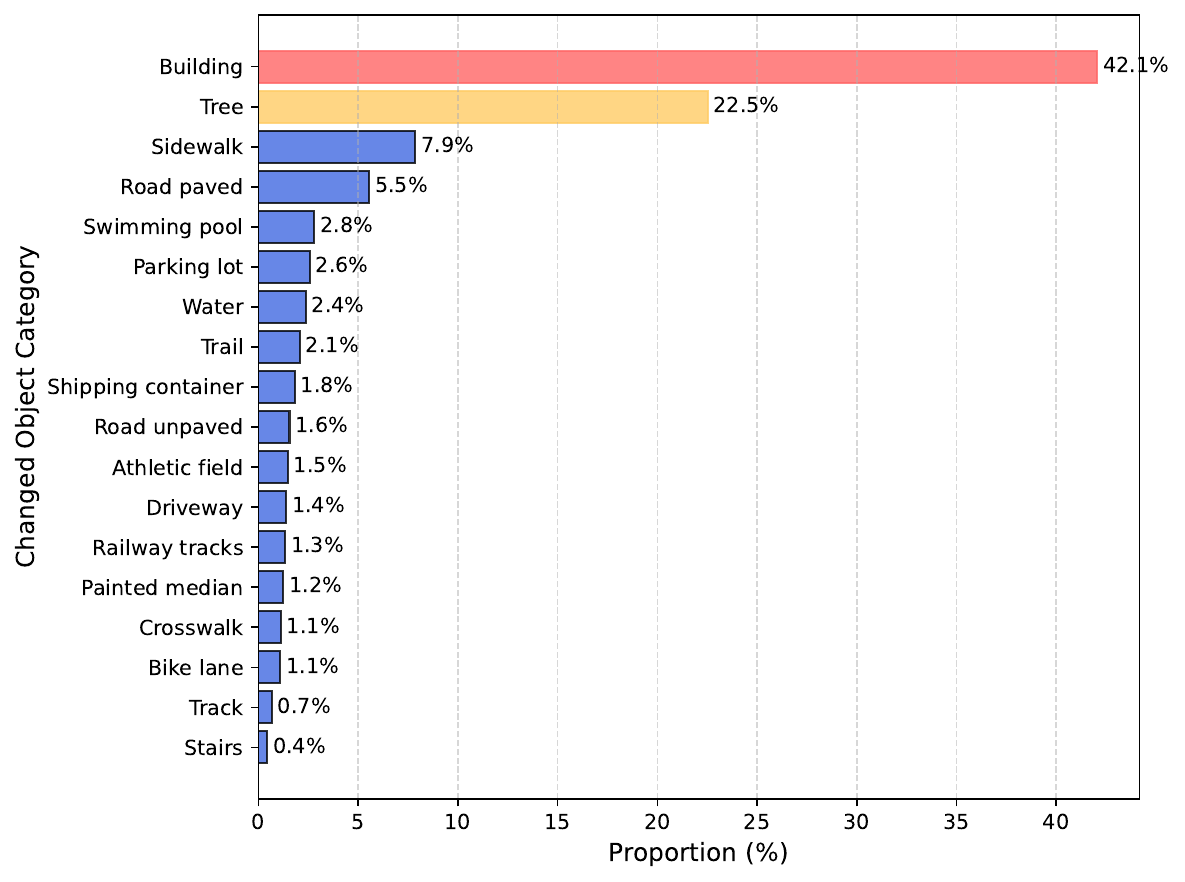}
    \caption{Class distribution of detected changes in the test set. 
    The majority of changes correspond to \textit{building} and \textit{tree} categories, 
    followed by smaller contributions from man-made and natural structures. 
    This distribution highlights the long-tail nature of the dataset.}
    \label{fig:class_test_dist}
\end{figure}

Table~\ref{tab:class_distribution} and Figure~\ref{fig:class_test_dist} present the distribution of detected changes across semantic categories in our generated dataset (test only). 
The results indicate a clear dominance of the \textit{building} and \textit{tree} classes, which together account for over 64\% of all detected changes. 
This imbalance aligns with the urban–suburban nature of the underlying LEVIR-CD imagery, where most structural and vegetation modifications occur.

While smaller categories such as \textit{roads}, \textit{sidewalks}, and \textit{parking lots} appear less frequently, they provide valuable diversity for model generalization. 
The long-tail behavior visible in Figure~\ref{fig:class_test_dist} emphasizes the need for adaptive sampling and weighting during training to ensure balanced learning across object types. 
Overall, this distribution validates the coverage of both man-made and natural change patterns in our dataset.

\begin{table}[!ht]
\scriptsize
\centering
\caption{Number of Detected Changes per Class in the Test Set. The proportion represents the ratio of each class relative to the total number of detected changes.}
\label{tab:class_distribution}
\begin{tabular}{l | c c}
\hline
\textbf{Changed Object Category} & \textbf{Number of Changes} & \textbf{Proportion (\%)} \\
\hline
\midrule
Building            & 53003 & 42.06 \\
Tree                & 28426 & 22.52 \\
Sidewalk            & 9920  & 07.86 \\
Road paved          & 7000  & 05.55 \\
Swimming pool       & 3539  & 02.80 \\
Parking lot         & 3253  & 02.58 \\
Water               & 2998  & 02.38 \\
Trail               & 2628  & 02.08 \\
Shipping container  & 2305  & 01.83 \\
Road unpaved        & 1964  & 01.56 \\
Athletic field      & 1872  & 01.48 \\
Driveway            & 1747  & 01.38 \\
Railway tracks      & 1686  & 01.34 \\
Painted median      & 1575  & 01.25 \\
Crosswalk           & 1429  & 01.13 \\
Bike lane           & 1383  & 01.10 \\
Track               & 857   & 00.68 \\
Stairs              & 546   & 00.43 \\
\midrule
\textbf{Total}      & \textbf{126131} & \textbf{100.00} \\
\hline
\end{tabular}
\end{table}

Table~\ref{tab:dataset_summary} provides a quantitative summary of our dataset after the automated filtering and validation stages. Out of the total 10,007 processed image pairs, 89,031 contain at least one verified change, while 37,100 are classified as \textit{no-change} samples. In total, the dataset includes 126,131 individual annotated changes, demonstrating both large-scale coverage and a significant volume of descriptive data. These statistics confirm the robustness of our automated pipeline in producing reliable, diverse examples suitable for high-capacity change detection and captioning tasks.

\begin{table}[ht]
\centering
\small
\caption{Summary of Processed Image Pairs}
\label{tab:dataset_summary}
\begin{tabular}{l | r}
\hline
\midrule
\textbf{Description} & \textbf{Count} \\
\hline
Rows with at least one change & 89,031 \\
Rows with no change & 37,100 \\
Total individual changes & 126,131 \\
\hline
\end{tabular}
\end{table}

\section{Human Agreement Across Dataset Creation Pipelines}

\begin{figure}[!ht]
    \centering
    \includegraphics[width=0.9\linewidth]{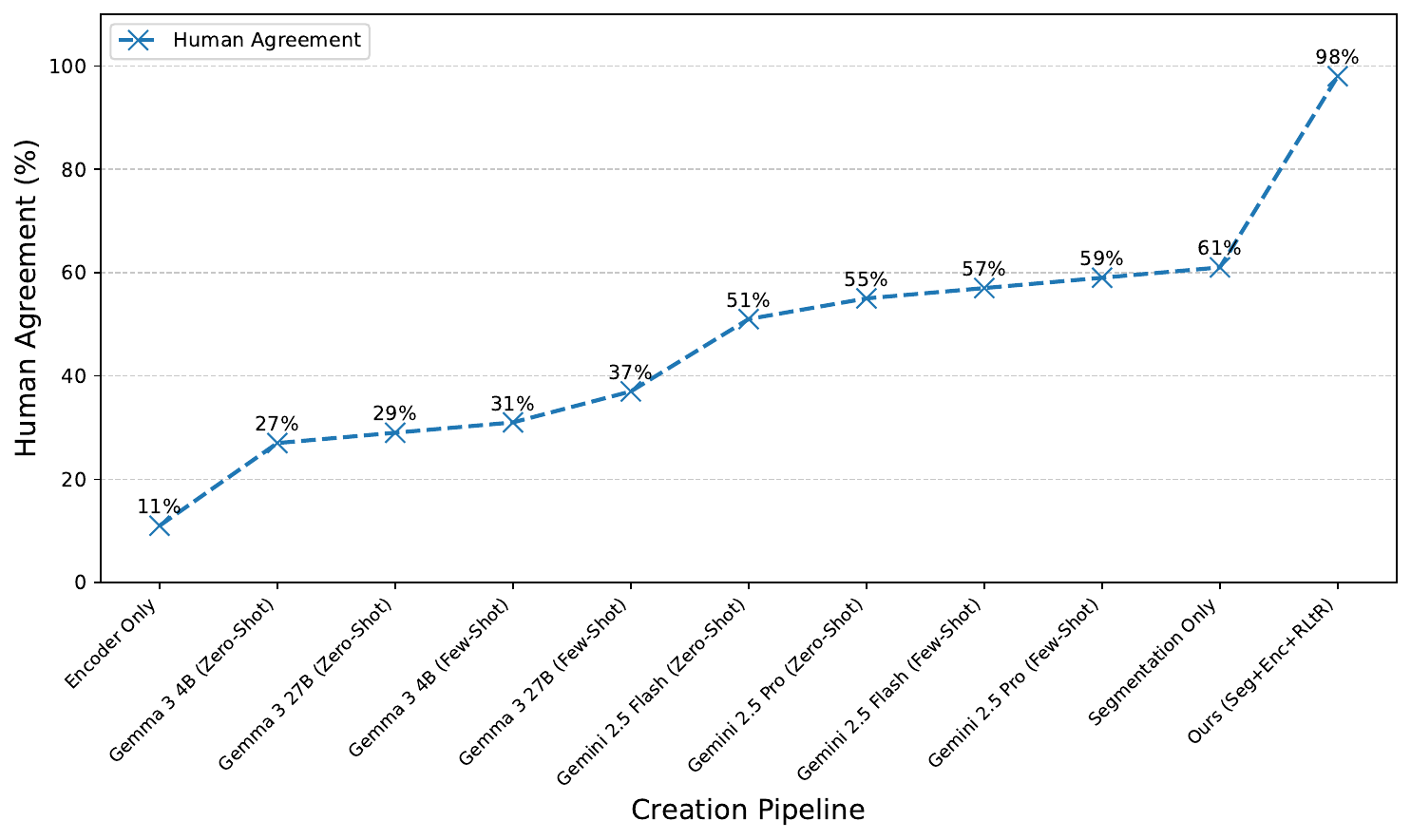}
    \caption{
    Human agreement across dataset creation pipelines.  
    Models are sorted in ascending order by agreement rate.  
    Few-shot variants consistently outperform their zero-shot counterparts, while our full method reaches near-perfect human alignment.
    }
    \label{fig:human_agreement}
\end{figure}

\noindent
As summarized in the results section and illustrated in Figure~\ref{fig:human_agreement}, 
our full pipeline (\textbf{Segmentation + Encoder + RLtR}) achieves the highest human agreement of 98\%, 
substantially outperforming all other baselines.  
Encoder-only and zero-shot models show limited semantic consistency, while few-shot adaptation-based on cosine-similarity retrieval-improves alignment across all LLMs.  
These results highlight the critical role of both visual filtering and reinforcement-based refinement in achieving human-level interpretability and reliability.

\section{Prompt Engineering}
\label{subsec:prompt_engineering}

\subsection{System Prompt for Gemma Filtering}
\label{subsubsec:gemma_prompt}

\begin{enumerate}

\item \textbf{Instruction:}
The model is prompted as an expert in satellite image interpretation to score a query patch based on the presence of a specific target class \(\{selected\_class\}\).
It receives reference examples and must assign a score from 1 to 5 according to visibility and clarity.

\begin{quote}
\texttt{``You are an expert in recognizing objects from satellite images. Your task is to score a query image patch from 1 to 5. You need to specify if \{selected\_class\} appears in the image. All images are satellite images. Return only the numerical score (1, 2, 3, 4, or 5).''}
\end{quote}

\item \textbf{Scoring Guide:}
The scoring criteria used for filtering are defined as follows:

\begin{quote}
\texttt{``5: There is definitely a \{selected\_class\} in the last image. The object's shape, shadow, and features are clearly visible from above.}\\
\texttt{4: Very likely that the image contains a \{selected\_class\}. Features are mostly clear.}\\
\texttt{3: Probably the image contains a \{selected\_class\}, but visibility or details are ambiguous.}\\
\texttt{2: Unlikely that a \{selected\_class\} appears in the image.}\\
\texttt{1: Definitely does not contain a \{selected\_class\}.''}
\end{quote}

\item \textbf{Example Format:}
The model is shown several examples of reference and query images structured as:

\begin{quote}
\texttt{``Example (1): \{start\_of\_image\} Score = 5\\
Example (2): \{start\_of\_image\} Score = 3\\
...\\
Example (5): \{query\_image\} Score = ?''}
\end{quote}

\end{enumerate}

This systematic prompt design enforces consistent, interpretable visual filtering behavior, ensuring the model evaluates satellite imagery with quantitative reliability across all semantic classes.

All experiments in this section were conducted to analyze the effect of different prompt formulations on the Gemma-based satellite image filtering task. 
The goal is to evaluate how textual guidance impacts the model’s ability to accurately score the presence of a target semantic class within a given query image.

For each image sample \( I \), the model receives a textual instruction prompt \( T \) and returns a classification score prediction:
\begin{equation}
R = \mathcal{V}(T, I),
\end{equation}
where \( \mathcal{V} \) denotes the Gemma inference function parameterized by the given prompt.

To systematically evaluate the effect of context and structure, we experiment with progressively more informative prompt variants as described below.

\begin{enumerate}

\item \textbf{Instruction-only:}  
A minimal instruction without additional guidance.  
\begin{quote}
\texttt{``You are an expert in recognizing objects from satellite images. Your task is to score whether a given class appears in the last image. Return only the numerical score (1-5).''}
\end{quote}

\item \textbf{Score Guide:}  
Adds explicit definitions of scoring criteria to improve rating consistency.  
\begin{quote}
\texttt{``Score the image from 1 to 5, where 5 means the object is clearly visible, 3 indicates uncertainty, and 1 means it is definitely absent.''}
\end{quote}

\item \textbf{Image Examples:}  
Extends the prompt with several labeled examples of reference and query image pairs.  
\begin{quote}
\texttt{``Example (1): \{image\_1\} Score = 5.  
Example (2): \{image\_2\} Score = 3.  
Example (3): \{query\_image\} Score = ?''}
\end{quote}

\item \textbf{Combined (Final):}  
Combines both the scoring guide and reference examples for maximum clarity and contextual learning.  
\begin{quote}
\texttt{``You are an expert in satellite image interpretation.  
Rate whether the object class appears in the last image, using a score from 1 to 5.  
Follow the scoring guide:  
5 = Definitely visible;  
4 = Very likely visible;  
3 = Unclear;  
2 = Unlikely;  
1 = Definitely not visible.  
Example (1): \{image\_1\} Score = 5.  
Example (2): \{image\_2\} Score = 3.  
Example (3): \{query\_image\} Score = ?''}
\end{quote}

\end{enumerate}

To evaluate the effectiveness of each Gemma prompt variant, we employed a human evaluator to assess the model’s predictions on a set of 100 satellite image pairs. 
All pre-processing stages-including semantic segmentation, connected component extraction, and image-text encoding-were held constant to isolate the effect of the prompt itself.

\begin{figure}[!ht]
\centering
\includegraphics[width=0.65\textwidth]{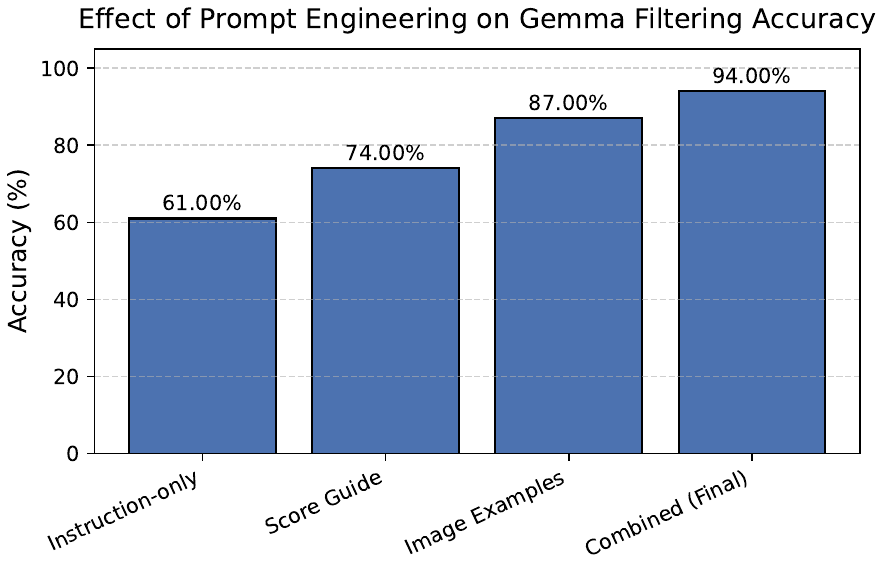}
\caption{Effect of prompt engineering on Gemma filtering accuracy.
Each bar represents the human agreement rate over 100 evaluation samples under fixed pre-processing conditions 
(segmentation, connected components, and image-text encoder).}
\label{fig:gemma_prompt}
\end{figure}

The results, illustrated in Figure~\ref{fig:gemma_prompt}, show that the simplest instruction-only prompt achieved a moderate agreement of 61\%. 
Introducing explicit scoring guidance improved interpretability and consistency, reaching 74\%. 
When example-based context was added, the accuracy further increased to 87\%, while the combined prompt, which integrates both scoring definitions and visual exemplars, achieved the highest human-aligned performance of \textbf{98\%}. 
These findings highlight that structured contextual prompts substantially enhance Gemma’s reasoning ability and scoring precision across visual categories.

\subsection{System Prompt for Gemini Question-Answer Creation}
\label{subsubsec:gemini_prompt}

To introduce controlled variability in question-answer generation, we set the model’s decoding temperature to \( T = 0.9 \). 
The temperature parameter regulates the degree of randomness during token sampling by scaling the logits \( z_i \) before applying the softmax normalization:
\begin{equation}
P(x_i) = \frac{\exp\left(\frac{z_i}{T}\right)}{\sum_j \exp\left(\frac{z_j}{T}\right)}.
\end{equation}
A higher temperature (\( T > 1 \)) increases output diversity by flattening the probability distribution, while a lower value (\( T < 1 \)) makes predictions more deterministic. 
Setting \( T = 0.9 \) balances creativity and consistency, ensuring natural linguistic variation without drifting from semantically valid question-answer pairs.

\begin{enumerate}

\item \textbf{Change-Present Instruction (\texttt{MCQ-Yes}):}  
The instruction guides the model to generate a multiple-choice question (MCQ) describing a visible change between two satellite images.  
Each question includes four options (\texttt{A-D}), where exactly one describes the actual change and the rest represent plausible but incorrect alternatives.

\begin{quote}
\texttt{``You are an expert in generating multiple-choice questions based on visual changes in satellite imagery.  
The image pair below shows a change related to \{cls\} in the red bounding box.  
Generate one multiple-choice question with 4 answer options (A, B, C, D) to describe this change.  
One option must correctly describe the change, and the other 3 must be incorrect.  
Ignore the red bounding box in your answers-it is only for you to understand the local change.  
Focus strictly on the change inside the red bounding box; all other differences are not correct.  
Format the output as:  
Question:\\nA.\\nB.\\nC.\\nD.\\nThe correct answer is:''}
\end{quote}

\item \textbf{No-Change Instruction (\texttt{MCQ-No}):}  
For cases where no visible change occurs, the model is instructed to generate a question where the correct answer explicitly identifies that there is \texttt{no change}, while the other options describe incorrect or misleading changes.

\begin{quote}
\texttt{``You are an expert in generating multiple-choice questions about visual comparisons in satellite imagery.  
The two images below show \textbf{no visible change}.  
Generate one multiple-choice question where the correct answer states that there was \textbf{no change},  
and the other three options must be incorrect changes (e.g., suggesting false changes).  
Format the output as:  
Question:\\nA.\\nB.\\nC.\\nD.\\nThe correct answer is:''}
\end{quote}

\end{enumerate}

This structured prompting design ensures balanced question–answer generation, maintaining semantic correctness for both change and no-change conditions while preventing overfitting to visual noise or irrelevant features.

\section{Robustness to the Choice of Segmentation Backbone}

To evaluate the sensitivity of the proposed filtering mechanism to the choice of segmentation backbone, we compare our custom segmentation model to two established baselines: SegFormer \citep{xie2021segformer}, a transformer-based semantic segmentation model, and Mask2Former \citep{cheng2022mask2former}, which jointly models pixel and mask representations. Both baselines were fine-tuned on the OpenEarthMap \citep{ye_openearthmap_2021}.
For each model, we generate segmentation masks on a representative subset of the dataset and compute the pairwise IoU between the resulting masks. Table~\ref{tab:segmentation_iou} reports the average IoU scores between the three segmentation models.
These results suggest that the proposed Best-of-$N$ filtering mechanism is largely insensitive to the specific segmentation backbone employed.

\begin{table}[!ht]
\centering
\small
\caption{Pairwise IoU between segmentation models.}
\label{tab:segmentation_iou}
\begin{tabular}{lccc}
\toprule
\textbf{Model} & \textbf{Ours} & \textbf{SegFormer} & \textbf{Mask2Former} \\
\midrule
Ours        & 1.00 & 0.94 & 0.96 \\
SegFormer   & 0.94 & 1.00 & 0.98 \\
Mask2Former & 0.96 & 0.98 & 1.00 \\
\bottomrule
\end{tabular}
\end{table}

\section{Filtering of the Discovery Pipeline}
\label{subsec:filtering_stats}

To better understand the behaviour of the unsupervised discovery pipeline,
we analyse the decisions made by the two sequential filtering stages:
(i) an \textit{image-text encoder} that performs initial semantic
screening of candidate masks, and (ii) an \textit{LLM judge} that evaluates
the remaining regions using retrieval-guided textual reasoning.
Only high-confidence pseudo-labels that pass both stages are used for
adaptation.

Table~\ref{tab:filter_stats} summarizes the filtering outcomes over the
full evaluation set. Out of total candidate regions produced by the segmentation model, $54.6\%$ were rejected by the image-text encoder.
The remaining ambiguous cases were forwarded to the LLM judge.
Among these, $23.3\%$ were accepted as valid semantic changes, while
$76.7\%$ were discarded as spurious or low-confidence detections.

\begin{table}[H]
\centering
\caption{Filtering statistics for the discovery pipeline.
The first stage applies an image-text encoder for semantic screening.
Candidates rejected at this stage are discarded.
Remaining regions are evaluated by an LLM judge, which accepts only
high-confidence pseudo-labels.}
\label{tab:filter_stats}
\begin{tabular}{lccc}
\toprule
\textbf{Stage} & \textbf{Rate} & \textbf{Decision} \\
\midrule
Total candidate regions & ---  & Generated by segmentation \\
\midrule
Rejected by image-text encoder & $54.6\%$ & Discarded \\
Forwarded to LLM judge &  $45.4\%$ & Evaluated \\
\midrule
Accepted by LLM judge & $23.3\%$ & Kept as pseudo-labels \\
Rejected by LLM judge  & $76.7\%$ & Discarded \\
\bottomrule
\end{tabular}
\end{table}

\section{Text Length and Linguistic Distribution Analysis}
\label{subsec:text_linguistic_dist}

\begin{figure*}[!ht]
    \centering
    \includegraphics[width=\textwidth]{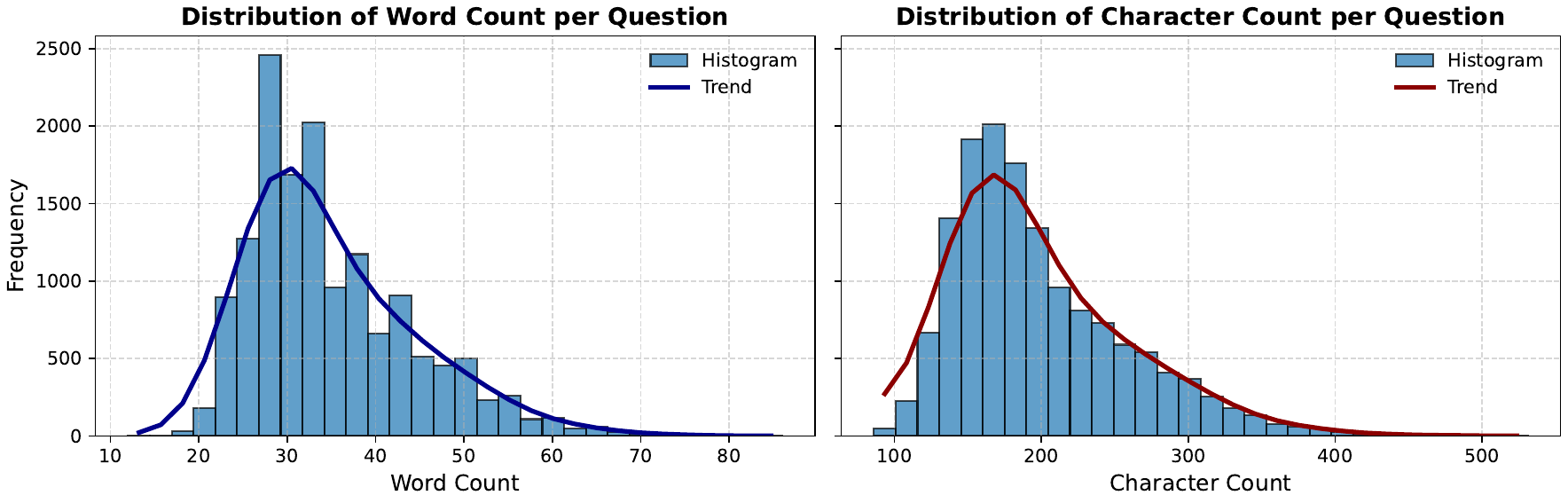}
    \caption{Distribution of textual complexity across generated answers.
    The left plot illustrates the distribution of word counts per question, while the right plot shows the corresponding character counts. Both follow approximately Gaussian-like trends centered around moderate lengths, indicating stable linguistic structure across the dataset.}
    \label{fig:text_dist}
\end{figure*}

\begin{figure*}[!ht]
    \centering
    \includegraphics[width=0.9\textwidth]{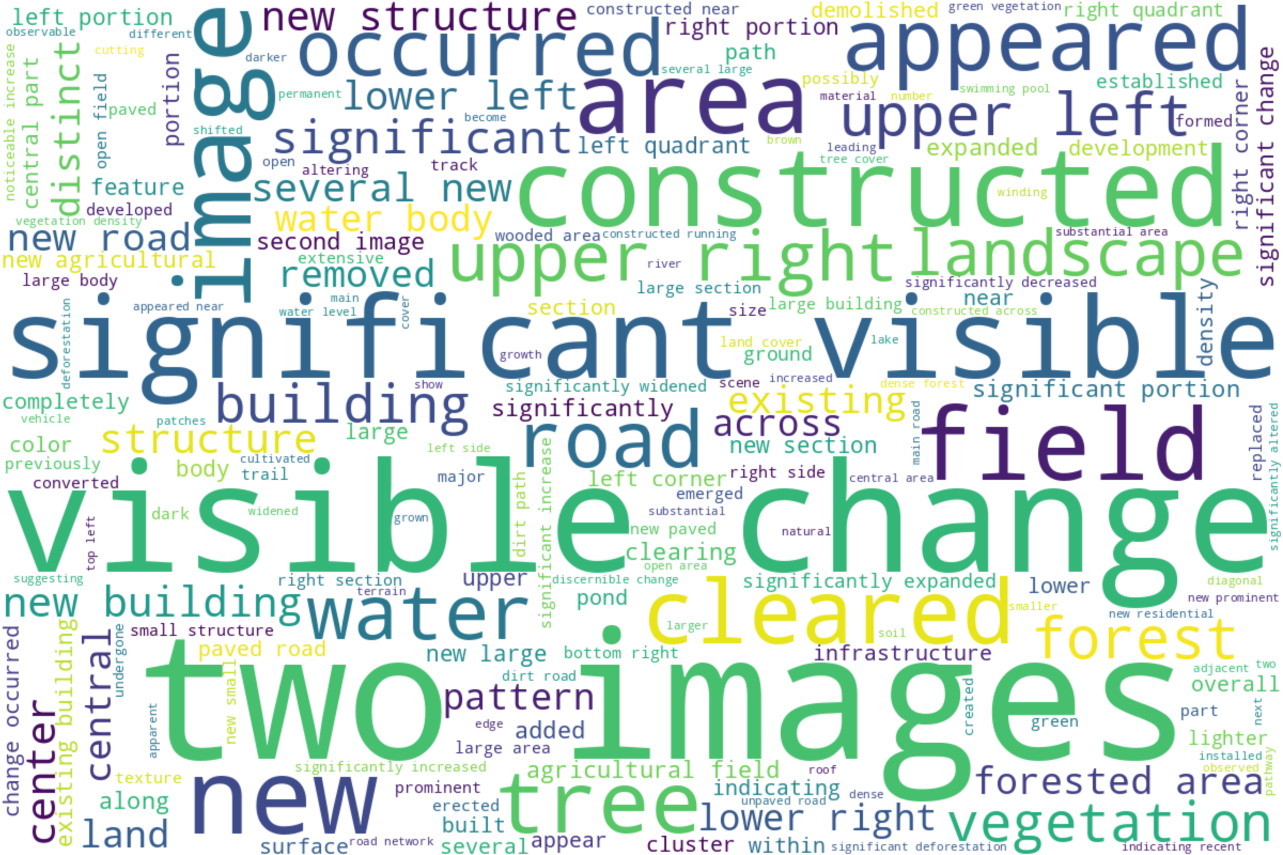}
    \caption{Word cloud visualization of the most frequent terms within the generated multiple-choice options. Larger words indicate higher frequency, highlighting recurring spatial and semantic patterns in the model outputs.}
    \label{fig:wordcloud}
\end{figure*}

To assess the textual and semantic properties of the generated multiple-choice questions, we jointly analyze the distributions of text length and linguistic frequency (Figures~\ref{fig:text_dist} and~\ref{fig:wordcloud}). The word and character count distributions exhibit smooth bell-shaped patterns centered around 40-50 words and 300-350 characters, respectively. This suggests that the generated content maintains a concise yet descriptive phrasing style, balancing informativeness and readability without excessive verbosity.

Complementary to this quantitative view, the word cloud highlights dominant lexical themes, with the most frequent terms including \textit{``visible,'' ``change,'' ``constructed,'' ``area,''} and \textit{``images''}. The frequent appearance of spatial descriptors such as \textit{``upper,'' ``right,'' ``left,''} and \textit{``central''} reflects the model’s focus on spatial reasoning within paired satellite imagery. Additionally, recurring terms like \textit{``new,'' ``cleared,''} and \textit{``forest''} capture consistent semantic structures tied to environmental transformation and construction-related contexts. Together, these findings confirm that the dataset preserves strong linguistic consistency and semantic relevance, supporting robust downstream model evaluation and interpretability.

\subsection{Compute Resources and Execution Environment}
All experiments reported in this work were conducted using a compute setup with \textbf{16 NVIDIA A100 GPUs}, each with \textbf{40GB of GPU memory}. This hardware configuration was used for the segmentation, candidate filtering, retrieval-augmented validation, and benchmark construction stages. For Gemini-based generation and evaluation, we used the \textbf{official Gemini API} accessed an authorized API key, rather than a self-hosted deployment. This setup ensured consistent access to the same model family used throughout the benchmark construction and evaluation pipeline.

\section{Qualitative Examples from the Dataset}
\label{sec:qualitative_examples}

To provide qualitative insight into the proposed dataset, we present representative examples of image pairs and their associated multiple-choice questions.
Each example consists of two temporally separated satellite images depicting a localized change, accompanied by a question with four answer options and a designated correct response.
These examples illustrate the diversity of structural and land-use changes captured in the dataset, including construction, demolition, vegetation removal, and infrastructure development.

The presented samples demonstrate the ability of the dataset to support fine-grained visual reasoning over subtle temporal changes, highlighting its suitability for training and evaluating multimodal models for remote sensing change understanding.

\begin{figure}[!ht]
    \centering
    \includegraphics[width=0.75\linewidth]{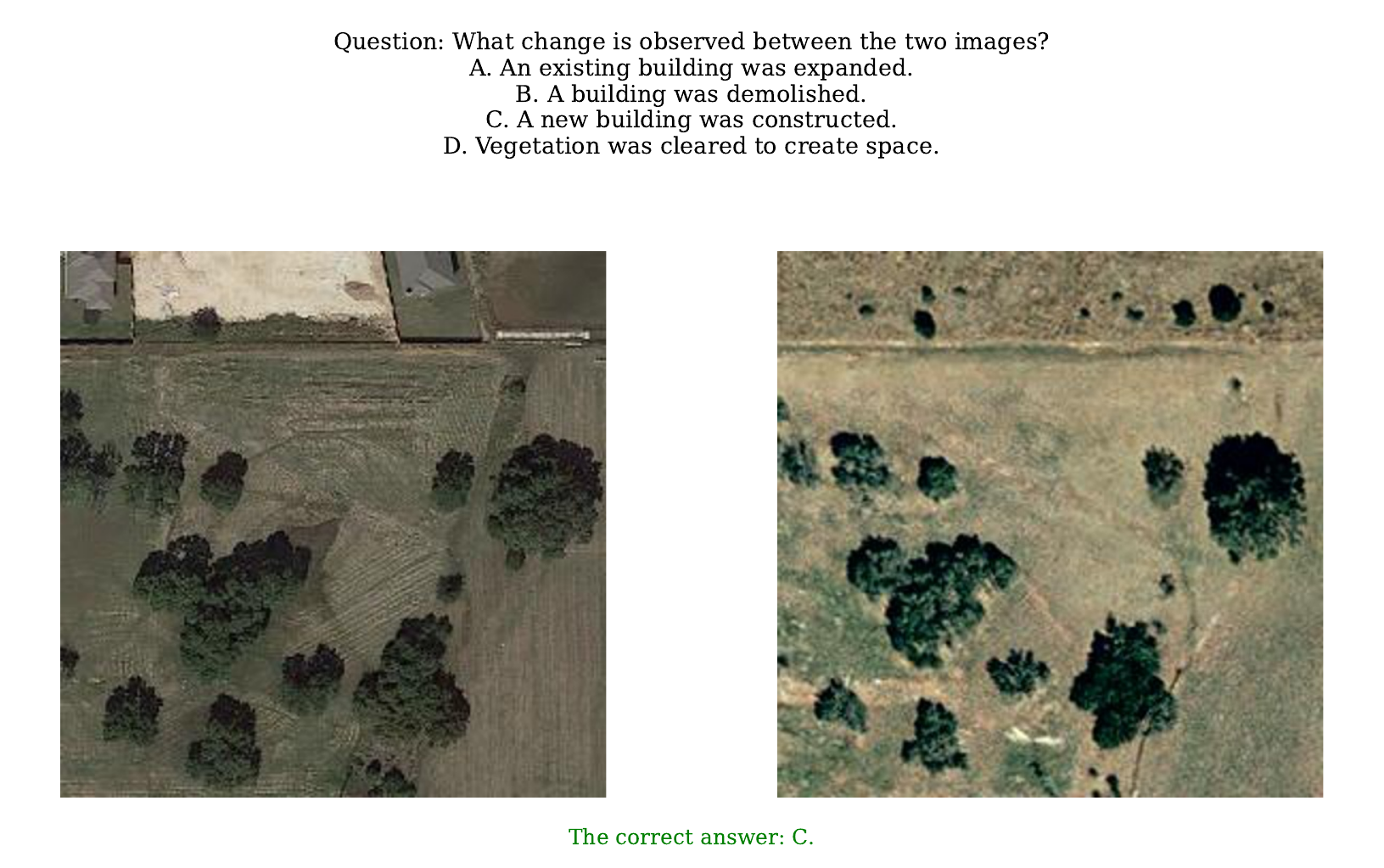}
    \label{fig:dataset_example}
\end{figure}

% \begin{figure}[!ht]
%     \centering
%     \includegraphics[width=0.75\linewidth]{figs/examples_1_0.pdf}
% \end{figure}

\begin{figure}[!ht]
    \centering
    \includegraphics[width=0.75\linewidth]{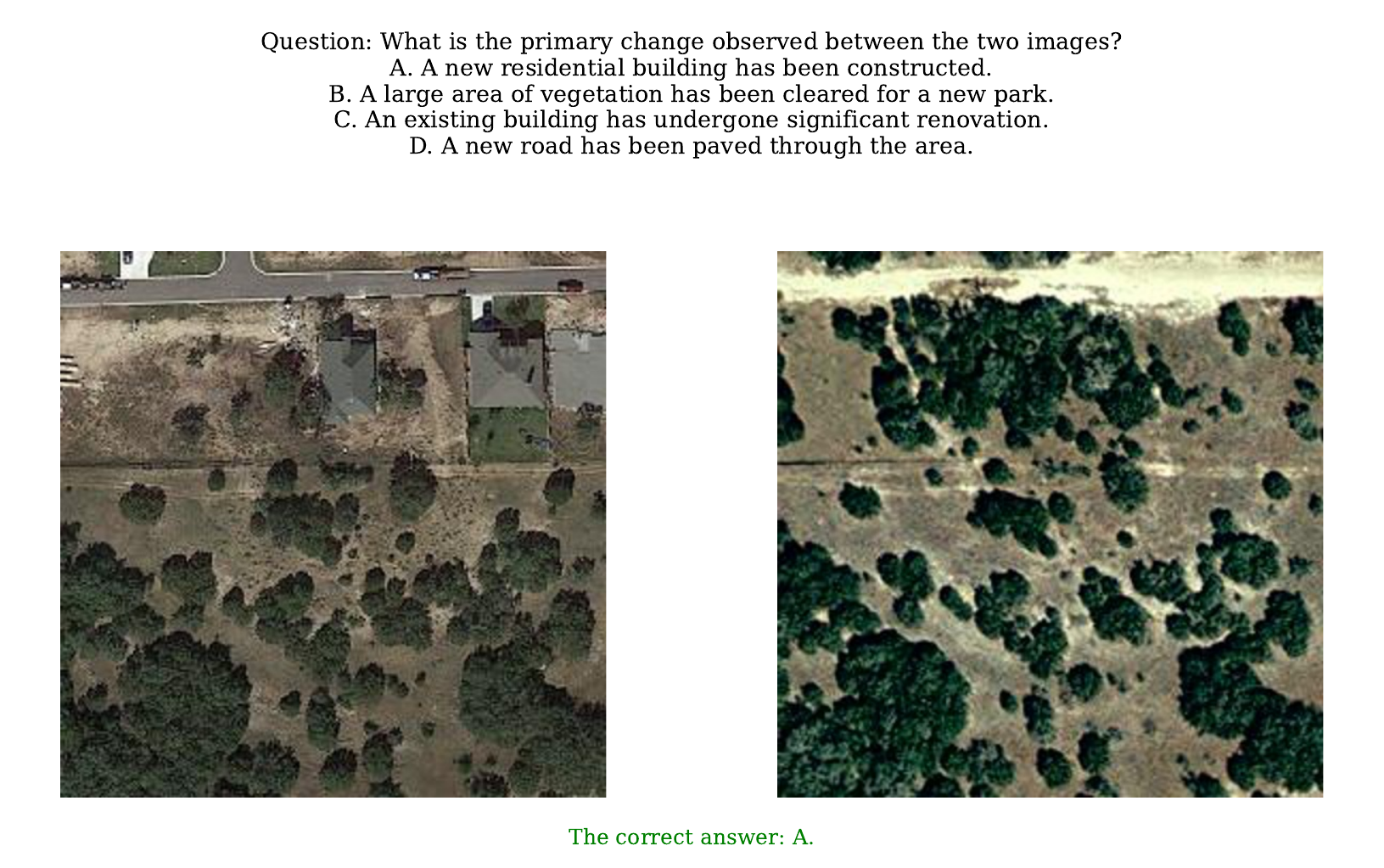}
\end{figure}

% \begin{figure}[!ht]
%     \centering
%     \includegraphics[width=0.75\linewidth]{figs/examples_2_6.pdf}
% \end{figure}

\begin{figure}[!ht]
    \centering
    \includegraphics[width=0.75\linewidth]{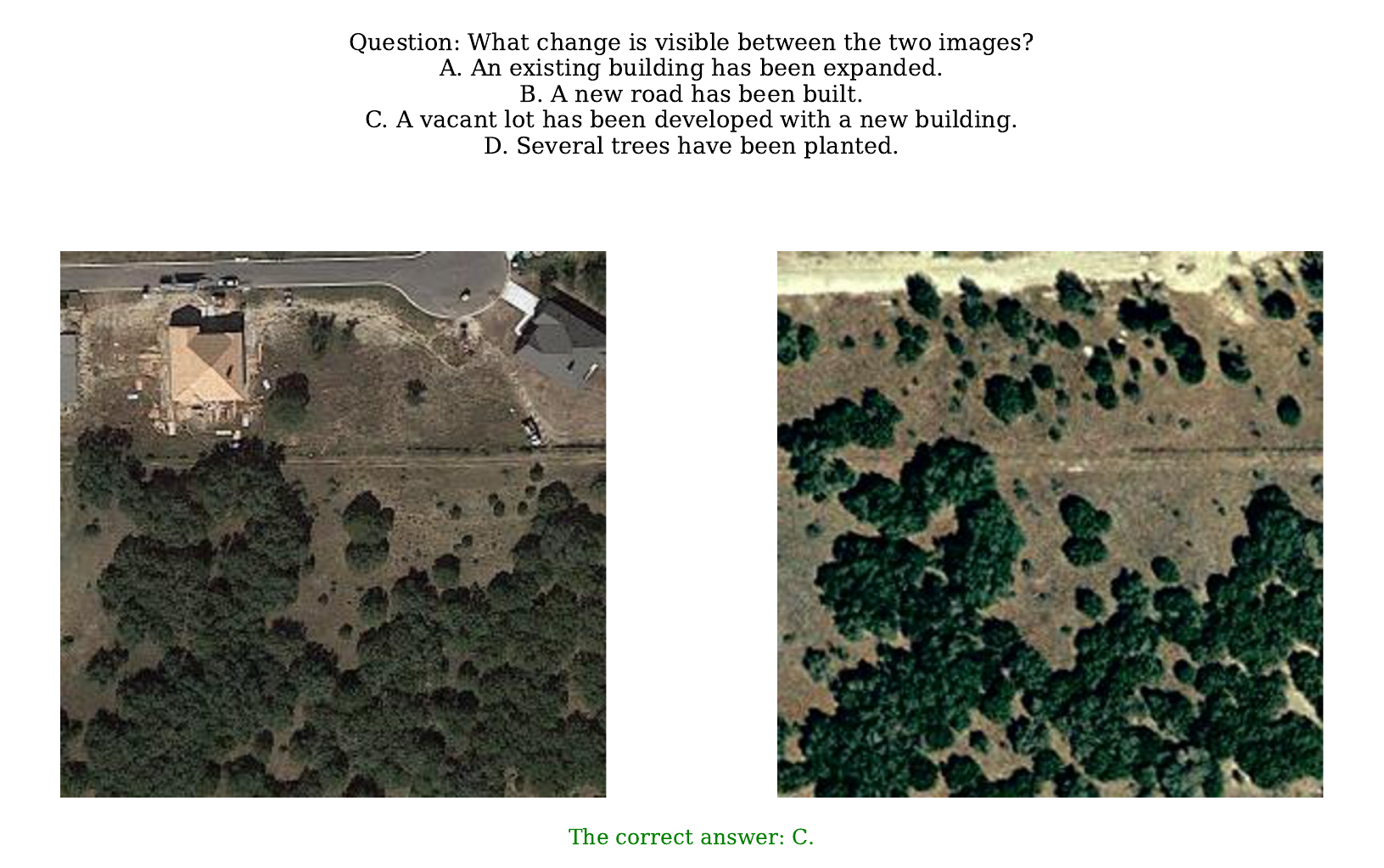}
\end{figure}

\begin{figure}[!ht]
    \centering
    \includegraphics[width=0.75\linewidth]{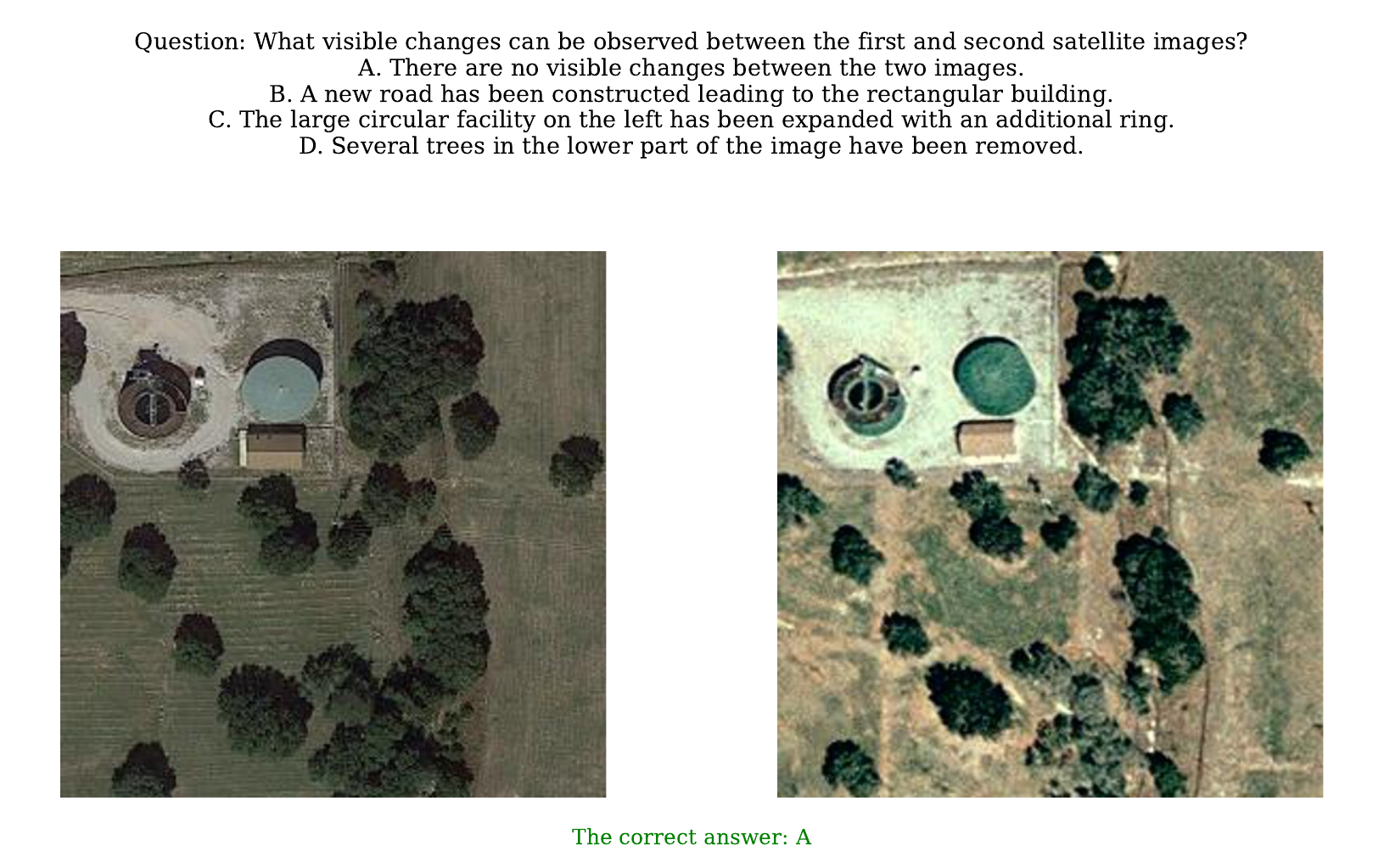}
\end{figure}

\begin{figure}[!ht]
    \centering
    \includegraphics[width=0.75\linewidth]{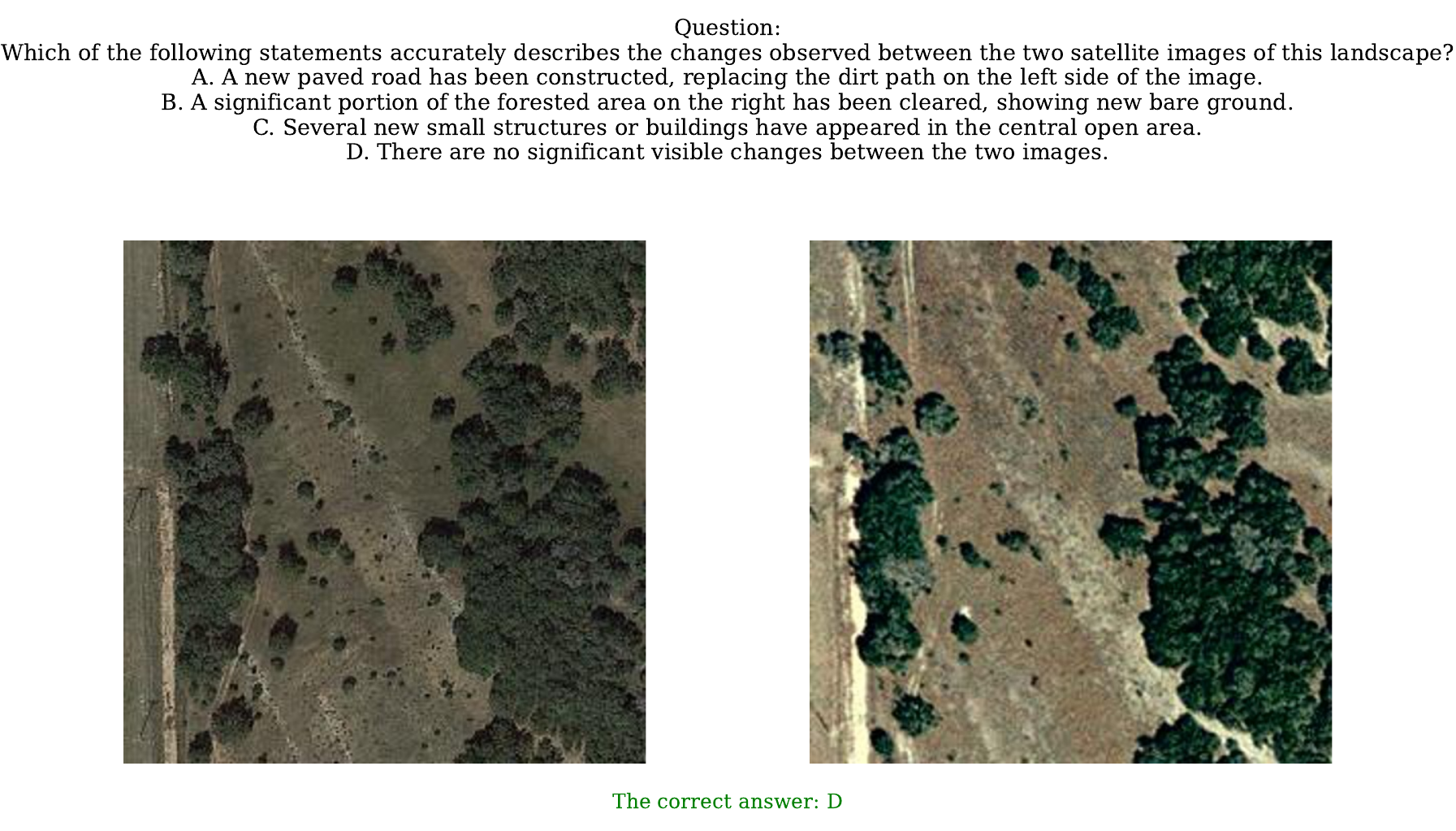}
\end{figure}

\begin{figure}[!ht]
    \centering
    \includegraphics[width=0.75\linewidth]{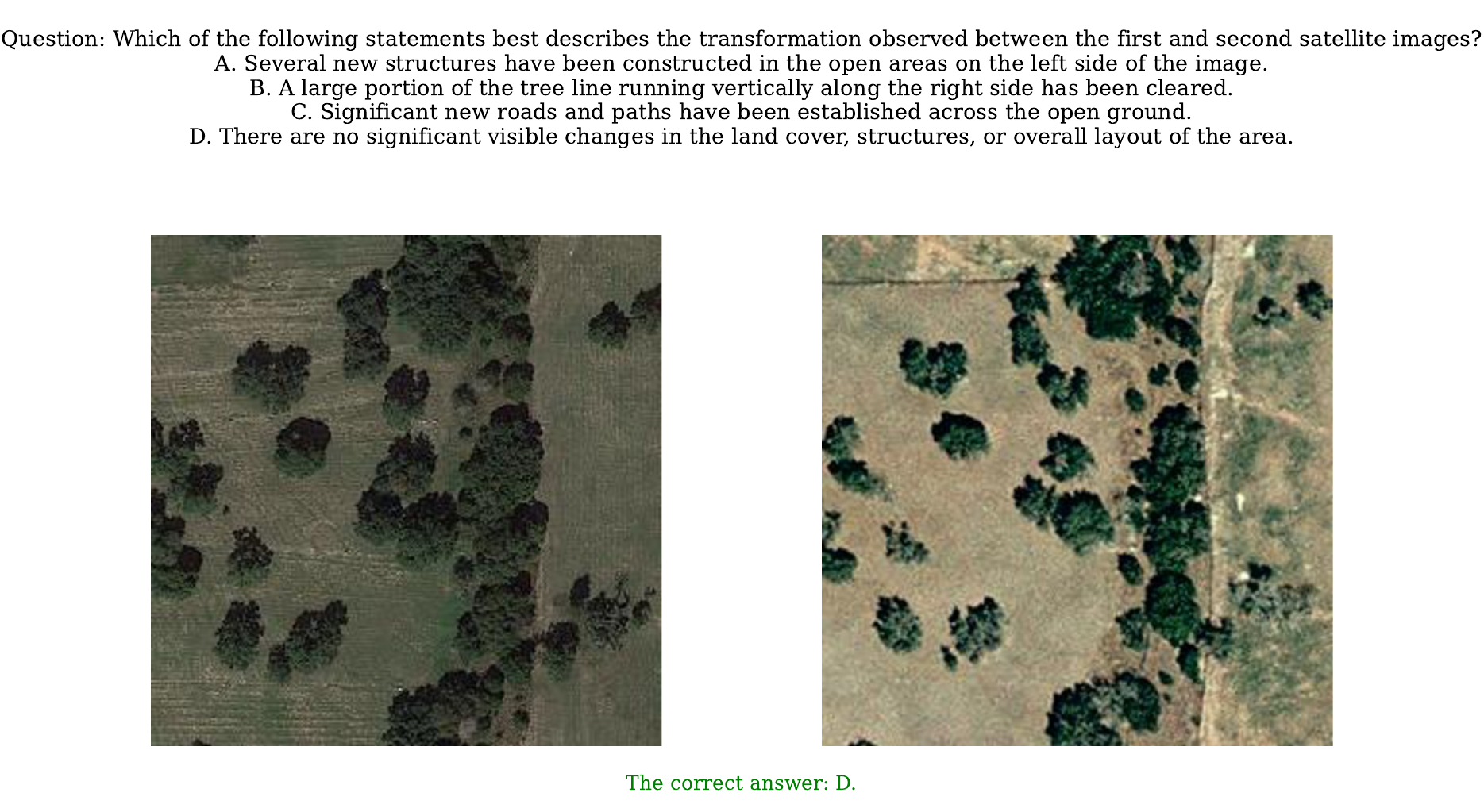}
\end{figure}

\begin{figure}[!ht]
    \centering
    \includegraphics[width=0.75\linewidth]{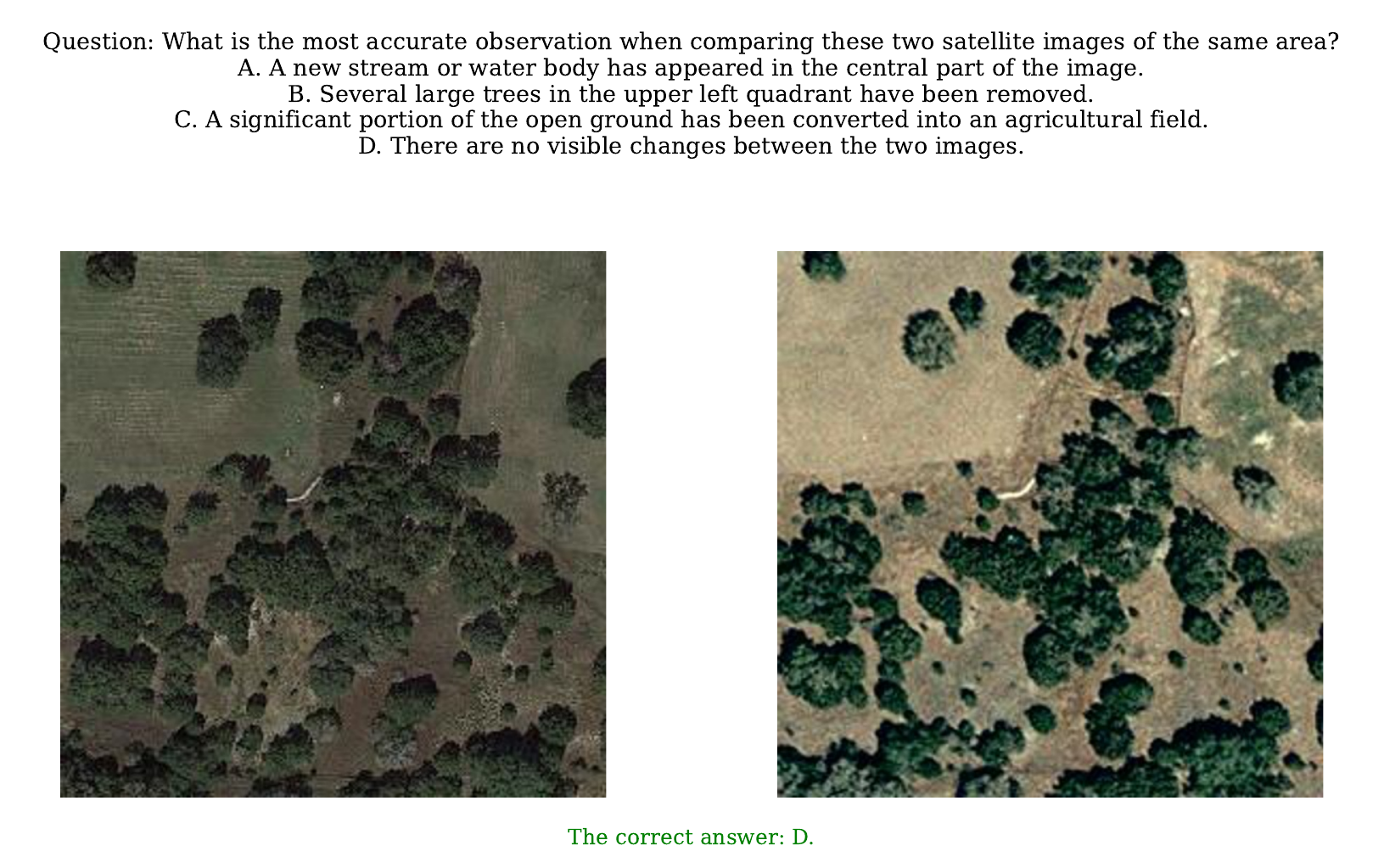}
\end{figure}

\begin{figure}[!ht]
    \centering
    \includegraphics[width=0.75\linewidth]{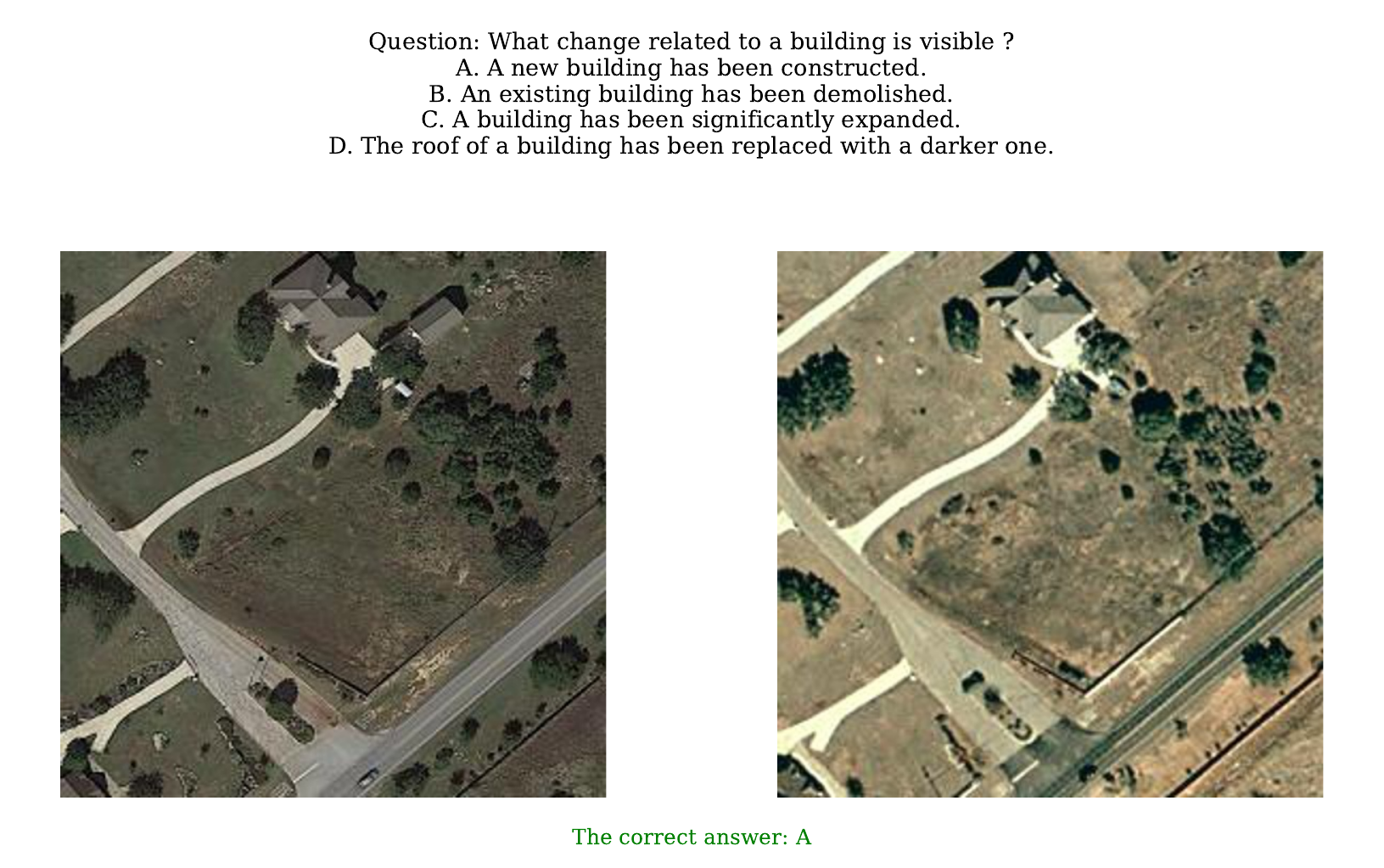}
\end{figure}

% \begin{figure}[!ht]
%     \centering
%     \includegraphics[width=0.75\linewidth]{figs/examples_20_0.pdf}
% \end{figure}

\begin{figure}[!ht]
    \centering
    \includegraphics[width=0.75\linewidth]{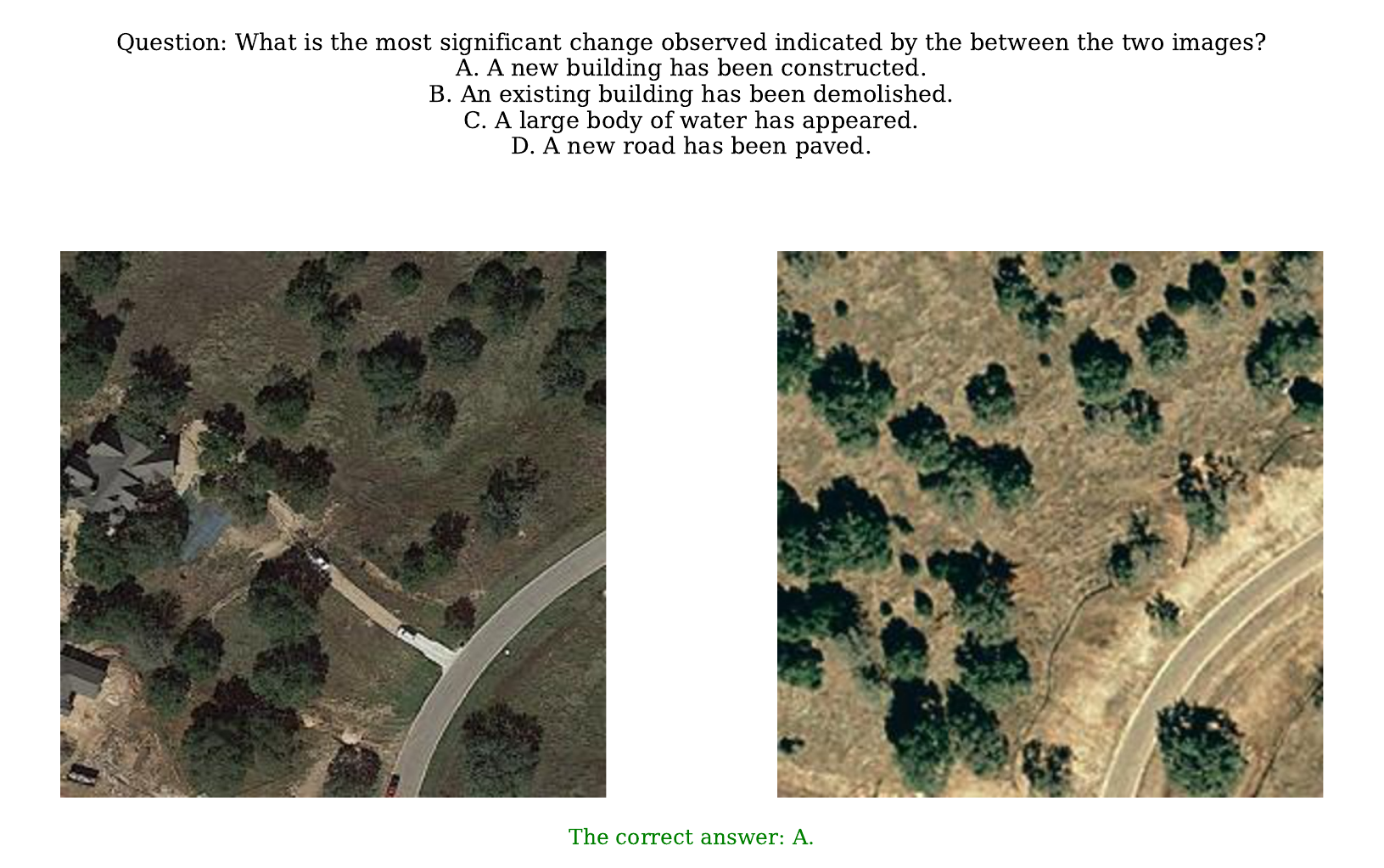}
\end{figure}

\begin{figure}[!ht]
    \centering
    \includegraphics[width=0.75\linewidth]{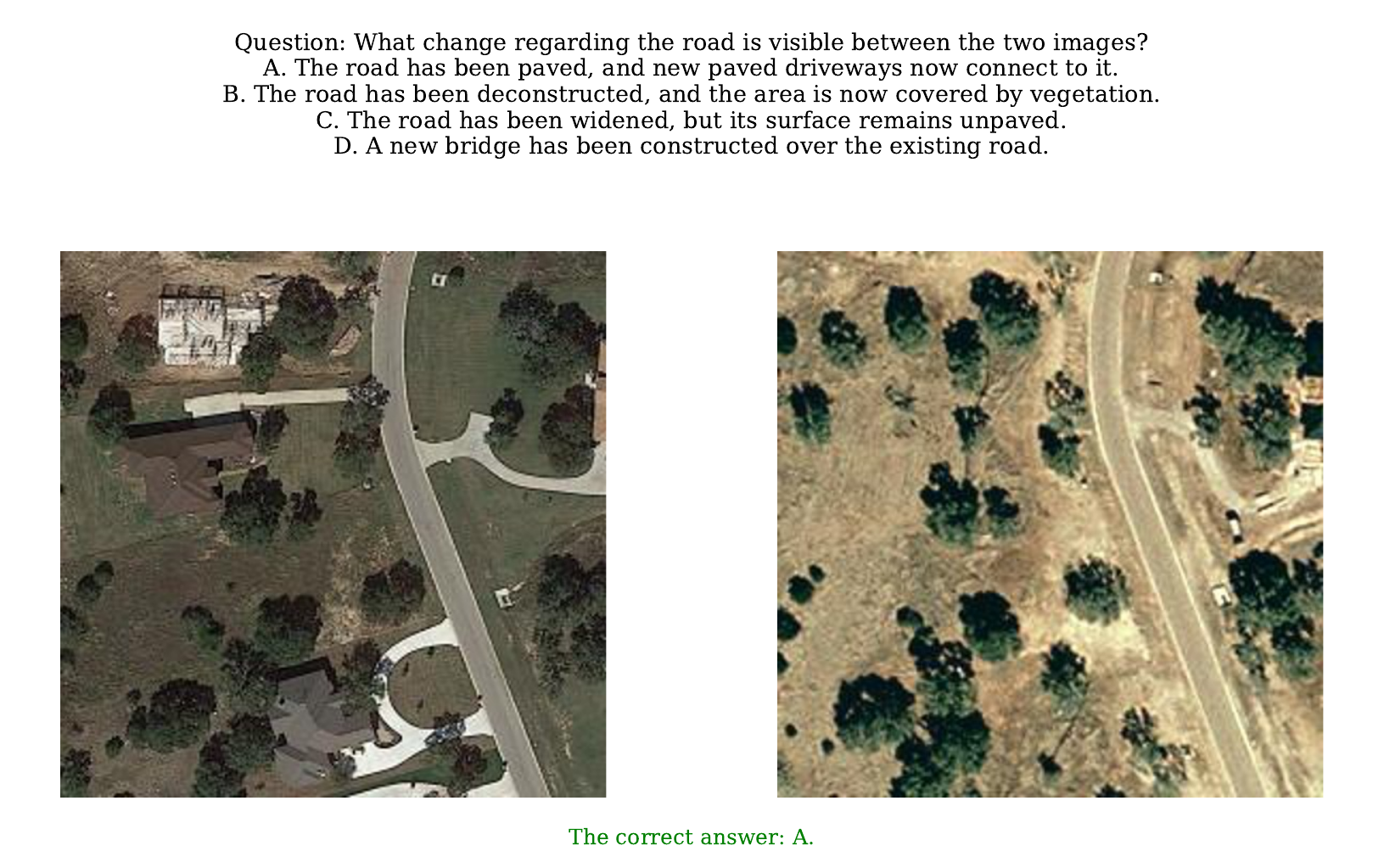}
\end{figure}

% \begin{figure}[!ht]
%     \centering
%     \includegraphics[width=0.75\linewidth]{figs/examples_28_2.pdf}
% \end{figure}

% \begin{figure}[!ht]
%     \centering
%     \includegraphics[width=0.75\linewidth]{figs/examples_30_11.pdf}
% \end{figure}

\begin{figure}[!ht]
    \centering
    \includegraphics[width=0.75\linewidth]{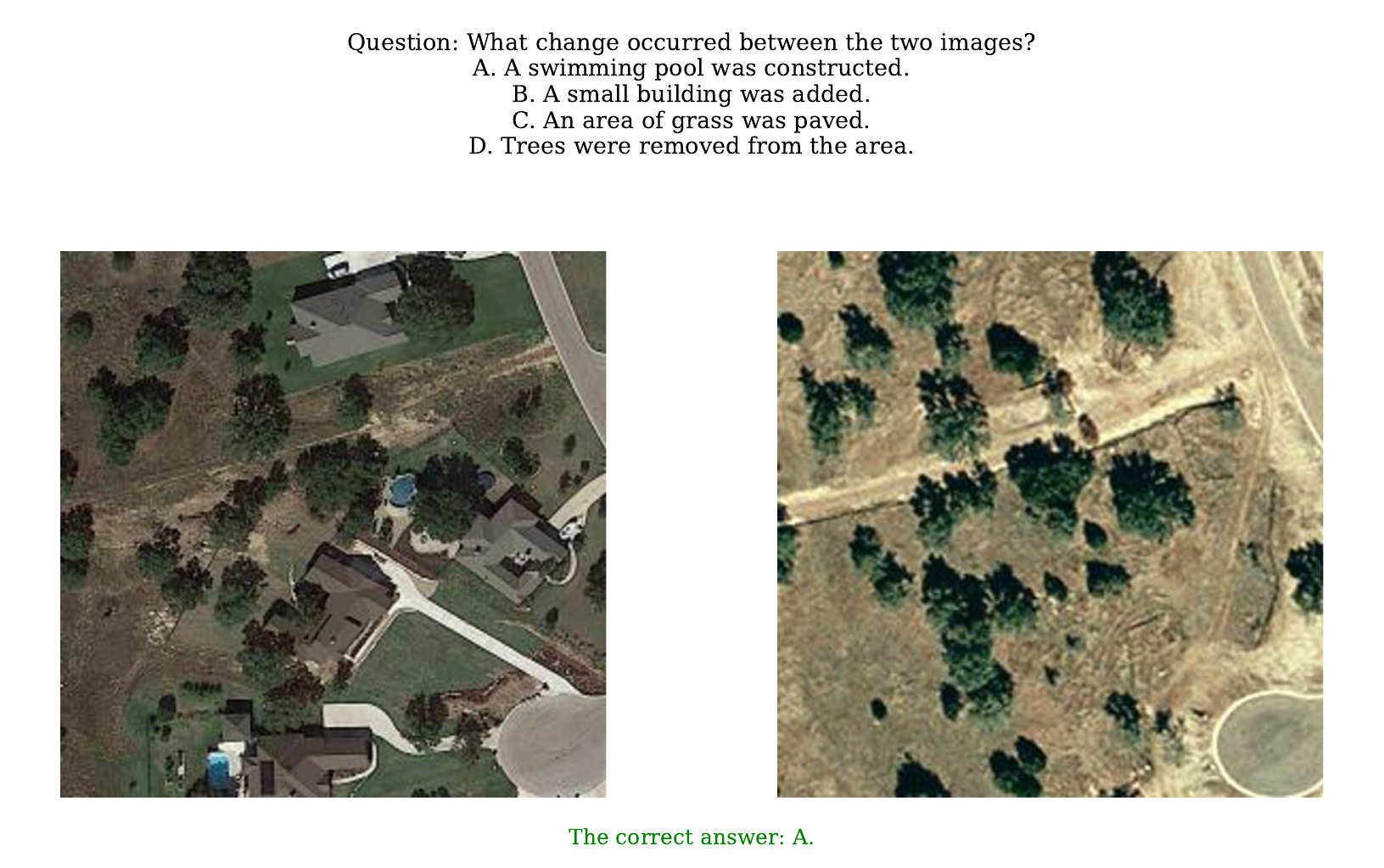}
\end{figure}

\begin{figure}[!ht]
    \centering
    \includegraphics[width=0.75\linewidth]{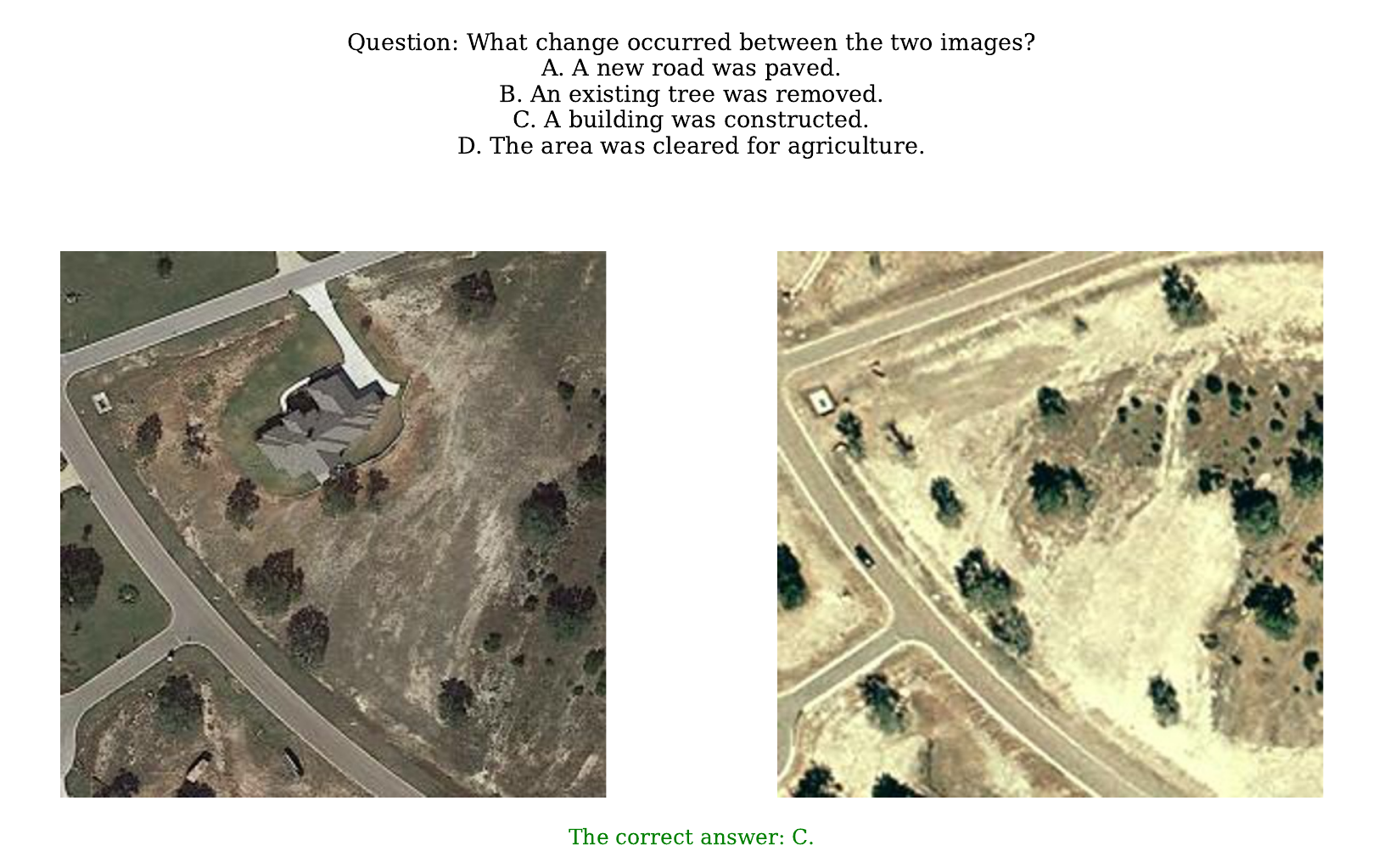}
\end{figure}

\begin{figure}[!ht]
    \centering
    \includegraphics[width=0.75\linewidth]{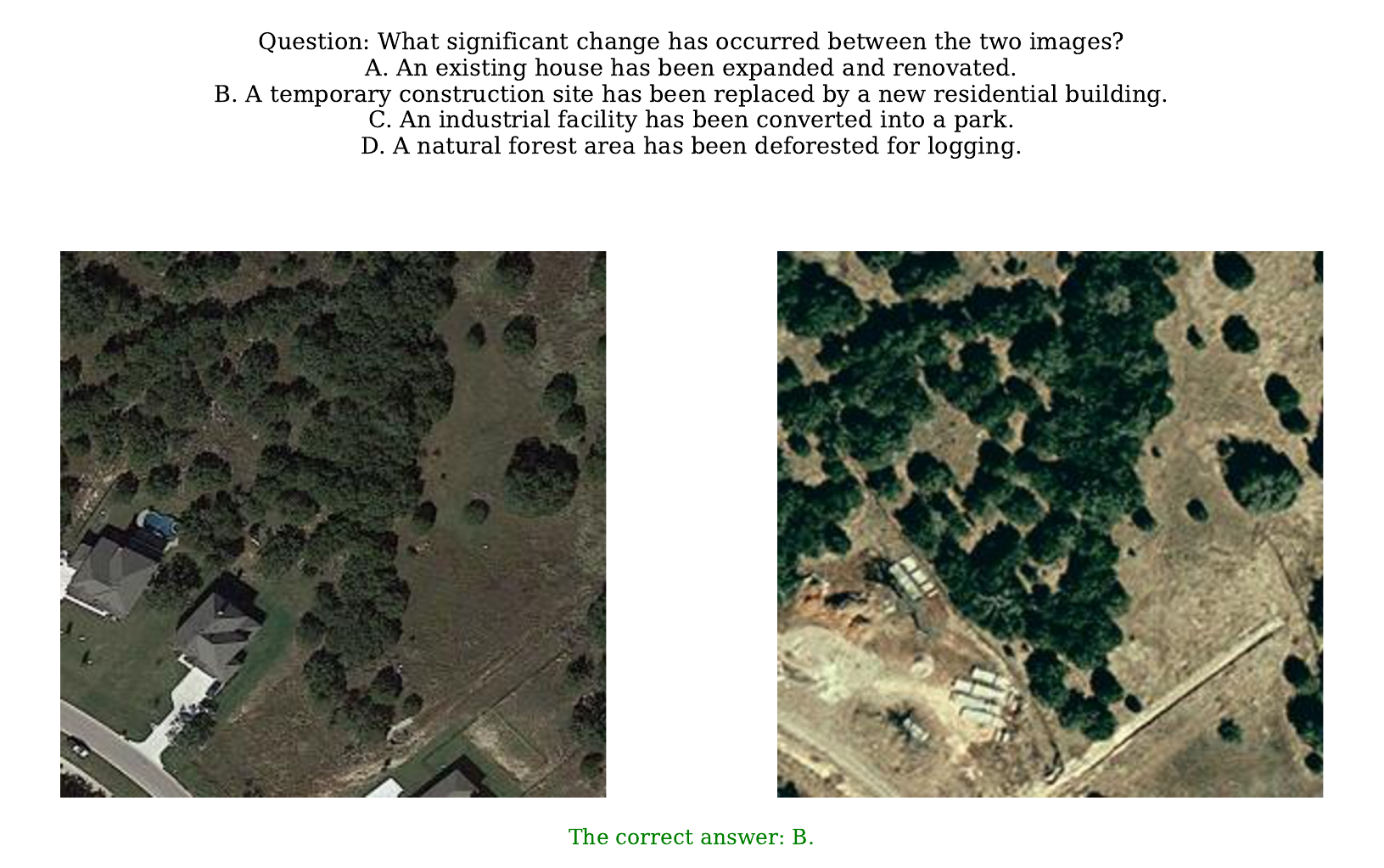}
\end{figure}

\begin{figure}[!ht]
    \centering
    \includegraphics[width=0.75\linewidth]{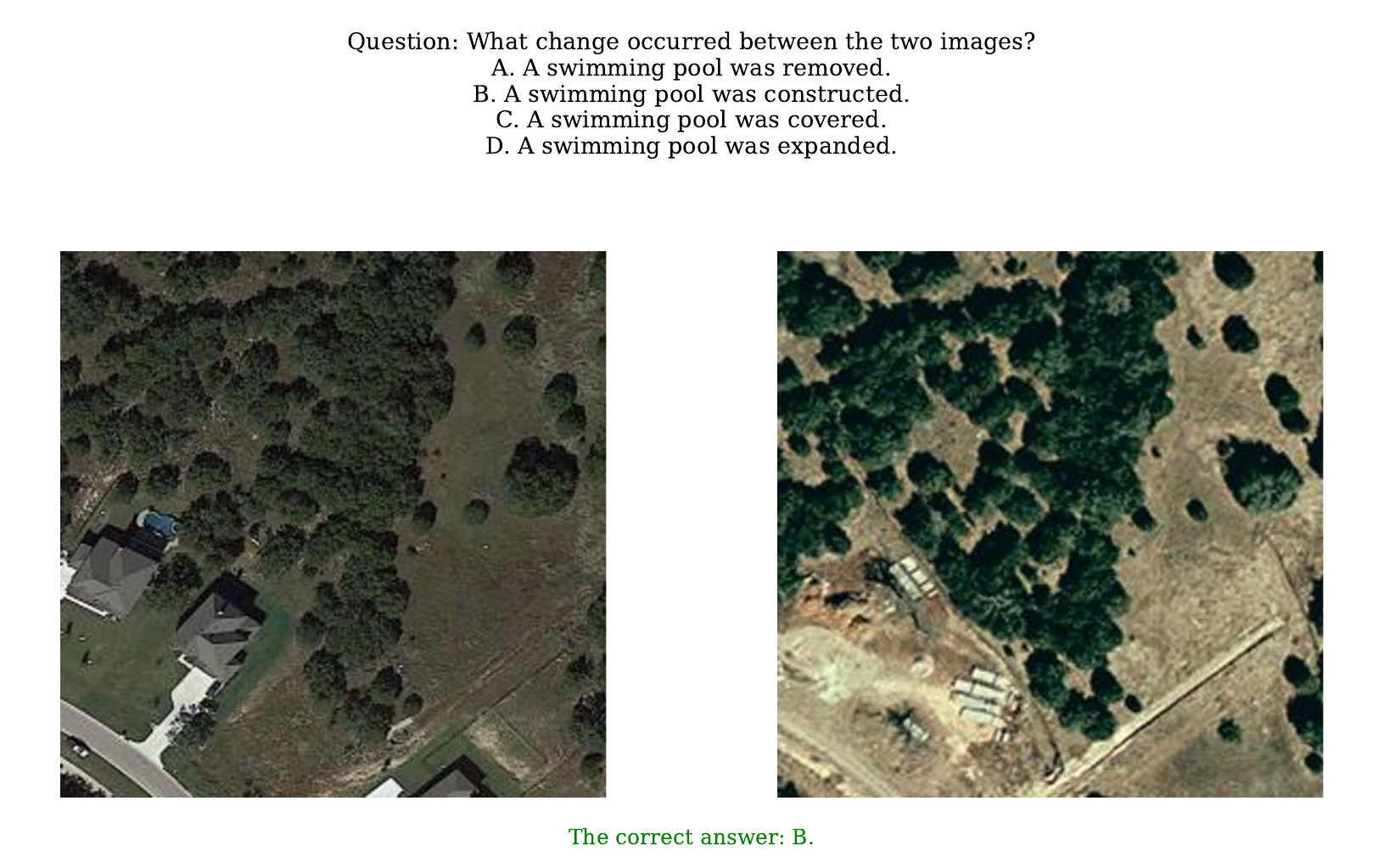}
\end{figure}

\begin{figure}[!ht]
    \centering
    \includegraphics[width=0.75\linewidth]{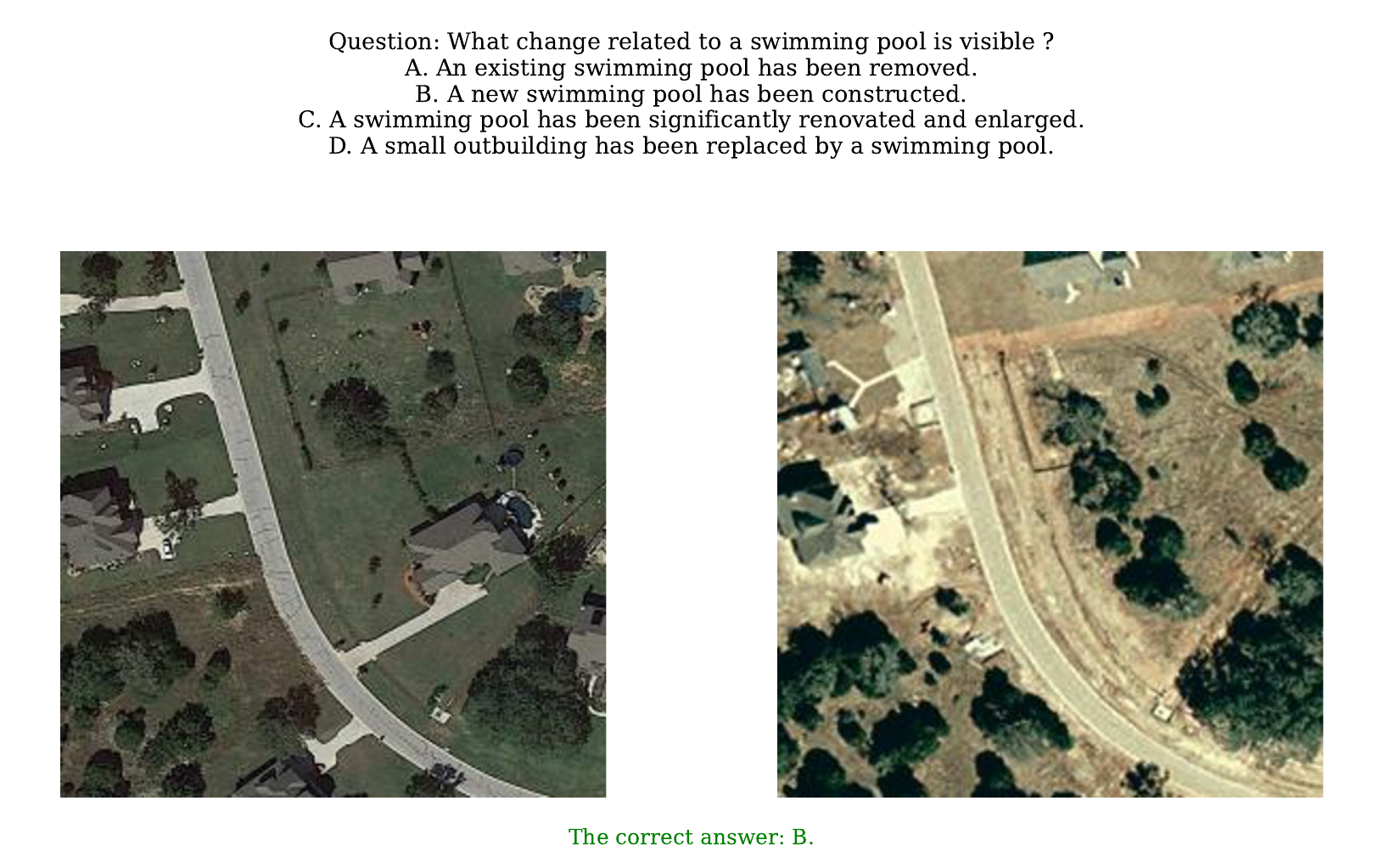}
\end{figure}

\clearpage

\section{Human Evaluation and Ethical Considerations}

We provide additional details on the human evaluation procedure used to validate the quality of the generated benchmark. Each evaluated example consisted of a pair of remote sensing images together with a generated question and its corresponding answer. The purpose of the evaluation was to assess whether the provided answer was correct with respect to the specific question being asked, rather than to exhaustively enumerate all changes present in the image pair.

\paragraph{Evaluators.}
Each example was independently reviewed by \textbf{three human evaluators}. All reported agreement statistics were computed from these three-way judgments.

\paragraph{Task instructions.}
Evaluators were given the following core guidance. They were told that the primary objective was to perform a high-fidelity verification of a specialized change-captioning dataset. For each task, they were shown a pair of remote sensing images alongside a question and answer generated by our pipeline, and were asked to indicate whether they \textit{agree} or \textit{disagree} with the provided answer. Evaluators were explicitly instructed that their role was \textbf{not} to judge whether the question was the most important possible question for the image pair, but only whether the proposed answer was correct for that specific question.

To avoid ambiguity, the instructions emphasized that each image pair may contain multiple visible changes, while the question may refer to only one localized object or semantic category. For example, if both a tree and a swimming pool changed, but the question referred only to the swimming pool, then the evaluator was asked to judge the answer only with respect to that object. The instructions also clarified that a change corresponds to a visibility difference between the two images, for example when an object is present in one image and absent in the other.

Evaluators were informed that the benchmark contains multiple question formats, including:
(i) binary yes/no questions that verify the presence or absence of a specific semantic change, and
(ii) multiple-choice questions that test whether the selected option correctly describes the observed change among several alternatives.

For examples labeled as \textit{no change}, evaluators were instructed to mark \textit{Disagree} only if they observed a visible change belonging to one of the predefined semantic categories used in the benchmark.

\paragraph{Annotation form.}
For each example, evaluators provided:
\begin{enumerate}
    \item an \textbf{Agree/Disagree} judgment indicating whether the answer correctly responded to the given question;
    \item an optional \textbf{improved alternative} in cases where the question or answer appeared unsatisfactory;
    \item a \textbf{difficulty score} from 1 to 3 reflecting how visually difficult the example was.
\end{enumerate}

The difficulty levels were defined as follows:
\begin{itemize}
    \item \textbf{1 (Very simple):} the change is clearly visible;
    \item \textbf{2 (Simple):} the change is visible, but requires a few seconds to localize or interpret;
    \item \textbf{3 (Hard):} the change is difficult to detect and may be partially obscured by shadows, occlusion, clutter, or challenging lighting conditions.
\end{itemize}

\paragraph{Presentation format.}
Evaluators were shown the paired remote sensing images together with the corresponding generated question and answer. The judgment was therefore made directly from the visual evidence and the textual query-answer pair for that example.

\paragraph{Agreement criterion.}
We adopt a strict unanimity-based definition of agreement. An example is counted as \textbf{agreement} only if \textbf{all three evaluators marked Agree}. If at least one evaluator marked \textit{Disagree}, the example was not counted as agreement. Formally, letting $a_i \in \{0,1\}$ denote the judgment of evaluator $i$, where $a_i=1$ corresponds to \textit{Agree}, the agreement indicator for an example is defined as
\begin{equation}
A = \mathbb{I}[a_1 = 1 \wedge a_2 = 1 \wedge a_3 = 1].
\end{equation}
The final human-agreement score is then computed as the fraction of evaluated examples satisfying this unanimity condition:
\begin{equation}
\text{HumanAgreement} = \frac{1}{N}\sum_{n=1}^{N} A_n.
\end{equation}

\paragraph{Interpretation.}
This unanimity criterion is intentionally conservative. It ensures that reported human-agreement rates reflect only examples for which there was complete evaluator consensus that the answer was correct for the given question. As a result, the reported scores should be interpreted as a strict lower-bound style estimate of annotation reliability rather than a lenient majority-vote metric.

To rigorously assess the semantic quality and accuracy of our generated dataset, we conducted a human evaluation study involving four external annotators. To ensure objectivity and mitigate potential bias, the evaluation was designed as a blind study. Annotators were presented with randomized samples from both our pipeline and baseline methods without knowledge of the specific source model or method associated with each sample. This ``blind'' setup ensured that the reported agreement scores reflect genuine visual-semantic alignment rather than any predisposition towards the proposed method.

% \newpage
% \input{checklist.tex}

\end{document}